\pdfoutput=1

\documentclass[11pt]{article}

\PassOptionsToPackage{table}{xcolor}
\usepackage[preprint]{acl}

\usepackage{times}
\usepackage{latexsym}
\usepackage{float}
\usepackage{graphicx}

\usepackage{booktabs}
\usepackage{amsmath}
\usepackage{amsfonts}

\usepackage{subcaption}
\usepackage{nccmath} 

\usepackage[T1]{fontenc}
\usepackage[utf8]{inputenc}
\usepackage{makecell}
\usepackage{pifont} 

\usepackage{microtype}

\usepackage{inconsolata}

\usepackage{multirow}
\usepackage{algorithm} 
\usepackage{algpseudocode} 

\usepackage{graphicx}
\usepackage{relsize}
\usepackage{tikz}

\title{LoSiA: Efficient High-Rank Fine-Tuning via \\
Subnet Localization and Optimization}

\author{
 \textbf{Xujia Wang\textsuperscript{}  }\ 
 \textbf{Yunjia Qi\textsuperscript{}  }\ 
 \textbf{Bin Xu\thanks{Corresponding author.}\textsuperscript{}}
\\
\\ 
 \textsuperscript{}Tsinghua University \\Department of Computer Science and Technology
\\
\texttt{wang-xj22@mails.tsinghua.edu.cn}
}

\begin{document}
\maketitle
\begin{abstract}

\looseness = -1
Parameter-Efficient Fine-Tuning (PEFT) methods, such as LoRA, significantly reduce the number of trainable parameters by introducing low-rank decomposition matrices. 
However, existing methods perform extensive matrix multiplications in domain specialization tasks, resulting in computational inefficiency and sub-optimal fine-tuning performance.
Hence, we propose LoSiA\footnote{The source code is released at \url{https://github.com/KlozeWang/LoSiA}. } (\textbf{Lo}w-Resources \textbf{S}ubnet \textbf{I}ntegration \textbf{A}daptation), an innovative method that dynamically localizes and optimizes critical parameters during the training process. Specifically, it identifies a sub-network using gradient sparsity analysis and optimizes it as the trainable target. 
This design enables effective high-rank adaptation by updating only the sub-network parameters, reducing the additional matrix multiplication.
We also present LoSiA-Pro, a faster implementation of LoSiA, which reduces training latency by about $27\%$ compared to LoRA. 
Extensive evaluations show that our method achieves minimal performance drop compared to full fine-tuning, while requiring the least training time across domain specialization and common-sense reasoning tasks.
Further analysis shows that LoSiA also reduces forgetting during continued training.

\end{abstract}

\section{Introduction}
\begin{figure}[htb]
    \centering
    \includegraphics[width=1.0\linewidth]{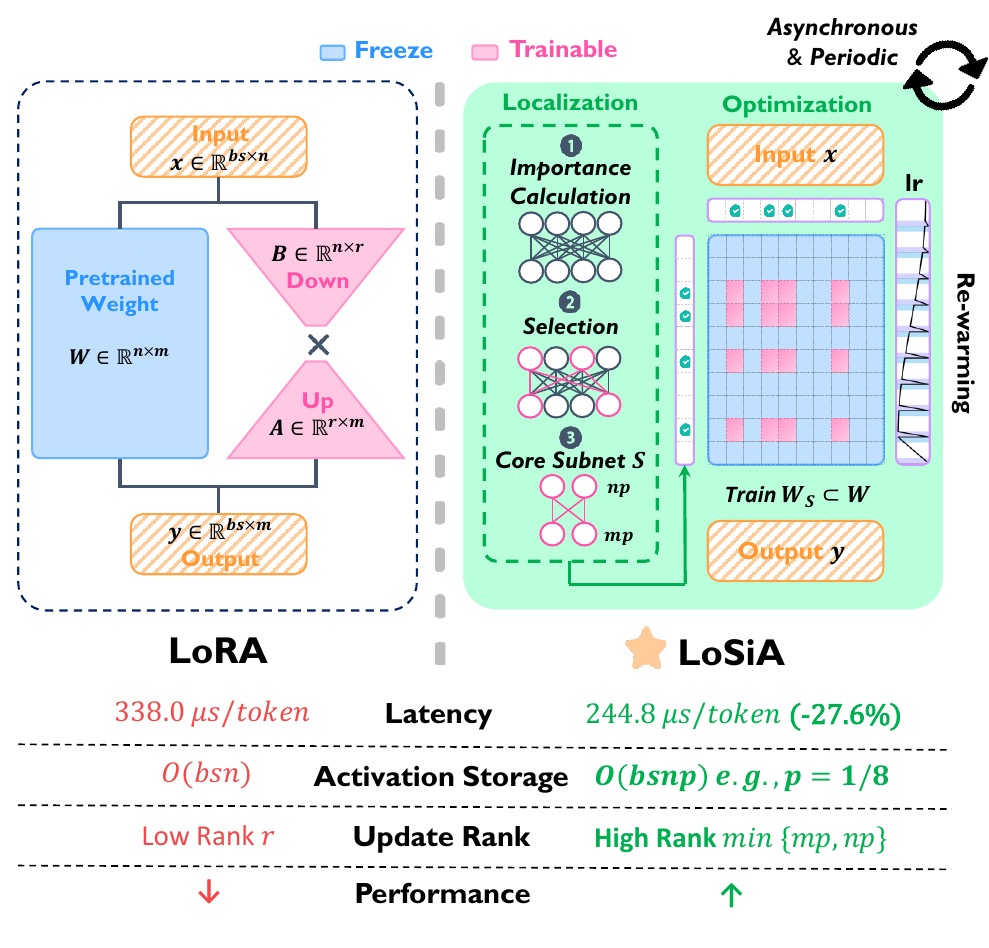}
    \caption{Overview of LoSiA. The method locates and optimizes core sub-network in asynchronous periods.}
    \label{fig:illu}
    \vspace{-17pt}
\end{figure}

Large language models, when fine-tuned via supervised learning, can be effectively adapted to downstream tasks such as mathematics~\cite{shao2024deepseekmath}, programming~\cite{hui2024qwen2}, and domain knowledge reasoning~\cite{wei2021finetuned}.
Although full parameter fine-tuning often yields the best performance, updating billions of parameters is computationally expensive and resource intensive. To address this, parameter-efficient fine-tuning (PEFT) updates only small amount of parameters to reduce GPU memory usage and communication overhead while maintaining performance comparable to full fine-tuning~\cite{houlsby2019parameter,ding2023parameter}.

Among PEFT approaches, LoRA~\cite{hu2022lora} has gained widespread adoption by introducing low-rank matrices to approximate full weight updates, producing competitive performance with significantly reduced computational and economic costs \cite{alpaca}.
Variants in the LoRA family further refine the method by biased fine-tuning modules \cite{zhu2024asymmetrylowrankadaptersfoundation,hayou2024loraefficientlowrank} or dimensions \cite{meng2024pissa} to accelerate convergence and achieve superior performance.
However, constrained by the low-rank assumption, these paradigms often struggle to balance model performance and efficiency, particularly in domain-specific tasks~\cite{yang2024lowrankadaptationfoundationmodels,ghosh2024closerlooklimitationsinstruction} and continual learning scenarios~\cite{shuttleworth2024loravsfinetuningillusion}. In such settings, low-rank configurations (e.g., $8$ or $16$) can lead to performance degradation and under-fitting~\cite{biderman2024loralearnsforgets}.
Although increasing the rank may mitigate these issues, it introduces additional memory consumption, extensive floating point operations, and risks of overfitting or convergence difficulties \cite{kalajdzievski2023rankstabilizationscalingfactor,borse2024fourafourierlowrank}. 
Recent studies have attempted to approximate high-rank updates by accumulating multiple low-rank components. However, these approaches still suffer from issues such as locally low-rank updates~\cite{lialin2023relorahighranktraininglowrank,meng2024periodiclorabreakinglowrankbottleneck} or increased computational complexity \cite{zhao2024galorememoryefficientllmtraining}. 
Therefore, while the low-rank assumption offers notable improvements in parameter efficiency, it also introduces inherent limitations.

The Lottery Ticket Hypothesis \cite{frankle2019lotterytickethypothesisfinding} suggests that dense neural networks contain trainable sub-networks capable of achieving comparable test accuracy. This prompts us to reconsider the PEFT roadmap and explore an alternative: \textbf{Can we identify and fine-tune such sub-networks within the backbone model to achieve high-quality adaptation more efficiently?}

To answer this question, we propose LoSiA (\textbf{Lo}w-Resources \textbf{S}ubnet \textbf{I}ntegration \textbf{A}daptation), a novel PEFT framework that dynamically localizes and optimizes critical sub-networks periodically, as illustrated in Figure \ref{fig:illu}. 
LoSiA asynchronously selects a core sub-network for each layer by calculating sensitivity-based importance scores and performing greedy selecting algorithms. Following localization, it fine-tunes the identified sub-network and applies a rewarming learning rate strategy to promote stable and robust training.
The design enables real-time high-rank updates without introducing additional matrix multiplication overhead, while significantly reducing training latency. Additionally, LoSiA does not introduce extra architectural components and only requires optimizer replacements for seamless deployment. Extensive experiments demonstrate its superior performance among PEFT baselines on domain-specific, commonsense reasoning tasks, while mitigating forgetting in continue learning. We also propose LoSiA-Pro, a more refined equivalent implementation of LoSiA, which significantly reduces the activation storage and computational complexity in backward propagation. LoSiA-Pro speeds up training $1.38\times $ compared to LoRA and $2.68\times $ compared to DoRA.\looseness-1

In summary, our contributions are as follows.

(1) Innovatively, we incorporate \textbf{sub-network structure} into the field of parameter-efficient fine-tuning. We devise a periodic workflow with techniques that localize, optimize, and integrate sub-networks, thus flexibly capturing and adapting task-essential parameters.\looseness-1

(2) We propose \textbf{LoSiA}, a novel high performance PEFT approach that dynamically localizes and optimizes sub-networks. By eliminating redundant computation, we further propose \textbf{LoSiA-Pro}, a loss-less variant that markedly reduces training latency and GPU memory footprint. \looseness-1

(3) We conduct extensive \textbf{evaluations} across multiple models and benchmarks.  
LoSiA outperforms all advanced PEFT baselines on domain-specific and common-sense reasoning tasks, while also accelerating training $1.15\times$ compared to LoRA.  
Moreover, its efficient variant, LoSiA-Pro, achieves a further speedup of $1.38\times$.

\section{Related Work}
\paragraph{\mbox{Parameter-Efficient Fine-Tuning}}
Full param-eter fine-tuning (FFT) adapts pre-trained models to downstream tasks by updating all model parameters~\cite{wei2022finetunedlanguagemodelszeroshot}, yet incurs prohibitive computational overhead.  \looseness-1
In contrast, parameter-efficient fine-tuning (PEFT) methods update only a small subset of parameters, curbing training costs while sustaining competitive accuracy. LoRA \cite{hu2022lora} approximates parameter updates as the product of low-rank matrices, achieving promising performance in tasks such as instruction tuning \cite{ghosh2024closerlooklimitationsinstruction}. Enhanced variants such as PiSSA \cite{meng2024pissa} accelerate convergence by prioritizing dominant singular vectors, while DoRA \cite{liu2024dora} decomposes updates into directional and magnitude components for more effective fine-tuning. Other derivatives such as LoRA+ \cite{hayou2024lora+}, LoRA-GA \cite{wang2024loragalowrankadaptationgradient}, and LoRA-Dash \cite{si2025taskspecificdirectionsdefinitionexploration} refine the framework by directional or module biased optimization. \looseness-1

However, recent studies \cite{jiang2024mora,biderman2024loralearnsforgets,ghosh2024closerlooklimitationsinstruction} reveal that the low-rank bottleneck restricts effectiveness in knowledge-intensive domains (e.g., mathematics, coding). Advanced solutions adopt strategies such as: 1) Architectural modifications through MoE-based LoRA combinations \cite{zadouri2023pushing,li2024mixloraenhancinglargelanguage,wang2024maloramixtureasymmetriclowrank} for multitask learning scenarios; 2) High-rank fine-tuning via accumulated low-rank projections, such as ReLoRA \cite{lialin2023relorahighranktraininglowrank}, MoRA \cite{jiang2024morahighrankupdatingparameterefficient} and GaLore \cite{zhao2024galorememoryefficientllmtraining}  to enhance training effectiveness. However, these ameliorated approaches either inflate architectural complexity or compromise throughput. Rare methods simultaneously optimize performance, training latency, and implementation simplicity. \looseness-1

\looseness-1

\paragraph{Skill Localization and Pruning}
LLM pruning compresses networks by excising redundant or less critical parameters. Previous work demonstrates that sparse networks can play crucial roles \cite{frankle2019lotterytickethypothesisfinding,yao2025knowledgecircuitspretrainedtransformers}. \citet{panigrahi2023taskspecificskilllocalizationfinetuned} identifies critical parameters in fine-tuned LMs by optimizing masks of grafted models, but such methods require additional training time and data. Alternatively, gradient- and sensitivity-based metrics enable real-time identification of task-aware parameters \cite{molchanov2019importanceestimationneuralnetwork,sanh2020movementpruningadaptivesparsity,zhang2022platonpruninglargetransformer}. Recent advances extend these ideas to PEFT: \citet{zhang2023adaloraadaptivebudgetallocation} prunes LoRA trainable parameters, while KIF \citep{feng2024kif,feng2024taslcontinualdialogstate} integrates skill localization into continual-learning regimes.

\section{Method}
\paragraph{Definition} Consider a model $f_0: \mathcal{X}\to \mathcal{Y}$ trained on dataset $\mathcal{D}=\{B_i\}_{i=1}^N$, where each batch $B_i=\{(x_j,y_j)\}_{j=1}^{M}$ contains $M$ samples. Let $W$ denote the parameters and $\mathcal{L}$ the loss function. The neural sub-network $S$ in $f_0$ is represented as the tuple $S = (X_{S},Y_{S},W_{X_S, Y_S})$, comprising its input neurons $X_S$, output neurons $Y_S$ and neural connections $W_{X, Y}$. Training model $f_0$ on $\mathcal{D}$ with full parameters $P_0=W$ is compactly written as $f_0\xrightarrow[\mathcal{D}]{P_0} f$. We investigate the following question: \textit{Given a cardinality budget, can we efficiently identify a parameter subset $P \subset P_0$, such that $f_0\xrightarrow[\mathcal{D}]{P} f'$ minimizes the loss difference $\Delta \mathcal{L} =|\mathcal{L}(f',\mathcal{D}) - \mathcal{L}(f, \mathcal{D})|$?} \looseness-1

\subsection{Structure of Gradients}
\label{sec: grd}
Inspired by pruning techniques, we expect to minimize the mean squared error (MSE) $\mathcal{L}_{\text{MSE}}$ between subsequent models $f_{k}$ and $f'_{k}$, where they both are trained from the model $f_{k-1}$ with trainable parameters $P_0$ and $P$, respectively. For SGD optimizers, we derive the bound:
\begin{align}
    \scriptstyle
    \mathcal{L}_{MSE} & \leq \eta ^2\frac{\|1_{(i,j)\not \in P}\cdot \nabla_{W_0} \mathcal{L} (\mathcal{B}_k)\|^2_F\|x\|^2_F}{M}
\label{eq:mse}
\end{align}

For AdamW, an analogous bound $\mathcal{L}_{MSE}$ holds in terms of $\nabla W$ in most cases (Appendix \ref{sec: grdprf}). Thus, gradient magnitudes lay in $P$ provide an upper bound for the approximation error, while retaining parameters with the largest gradient magnitudes to adjust tightens the bound.

Ideally, selecting the $\text{Top-}K$ entries of $\nabla W$ is theoretically optimal, but storing and fine-tuning sparse matrices compromises the efficiency. Instead, we claim that a suitable selection pattern for $P$ corresponds to a structured subnet $S=(X_{S}, Y_{S}, W_{X_S, Y_{S}})$, i.e., all connections between input neuron set $X_{S}$ and output neuron set $Y_{S}$.

\begin{figure}[t]
    \centering
    \vspace{-10pt}
    \includegraphics[width=0.9\linewidth]{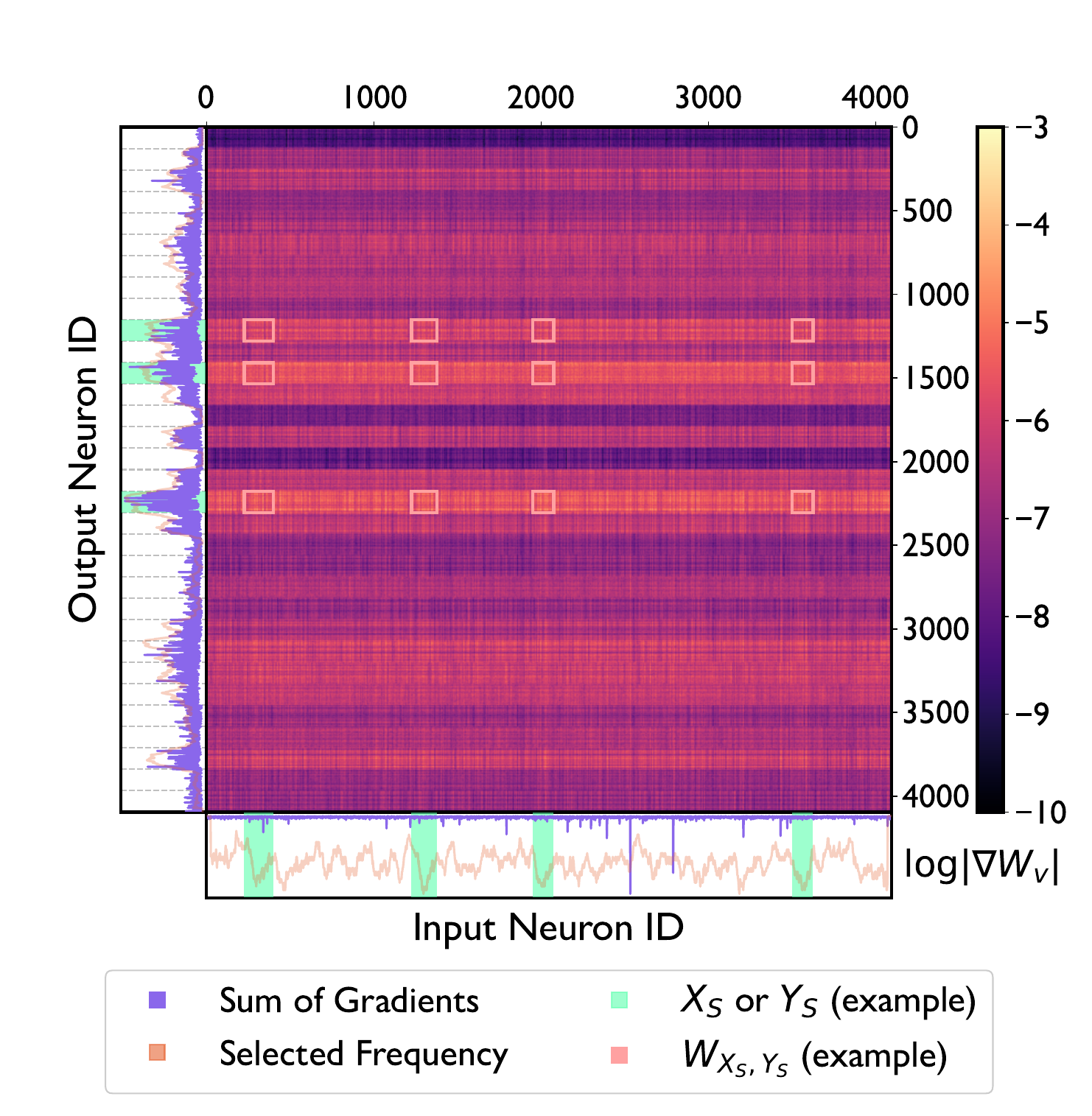}
    \caption{Gradient Magnitude Distribution of proj\_v. Large gradients follow a sparse subnet distribution.}
    \vspace{-15pt}
    \label{fig:grad}
\end{figure}

To validate this selection paradigm, Figure \ref{fig:grad} visualizes the gradient magnitude distributions in LLaMA-2 7B's \texttt{proj\_v} layer. Across the 32 attention heads, the gradient norms exhibit pronounced skewness and are highly correlated with the corresponding output neurons $Y_S$. On the other hand, a consistent set of input neurons $X_S$ (green markers, x-axis) contributes dominantly to all attention heads. The sparse pattern also holds in MLP layers (Appendix \ref{sec: grdobs}). Consequently, we restrict the fine-tuning space to subnet structures $S$ - termed the \textit{core subnet} - rather than the entire network.

\subsection{Subnet Localization}
To efficiently localize core subnets, an ideal algorithm should satisfy three key requirements: \ \ 1) Efficiency: no extra data or heavy computation. 2) Lightweight: negligible GPU memory overhead. 3) Dynamic Awareness: enable on-the-fly localization throughout the training process. Although existing LLM-pruning methods have achieved impressive compression ratios, they still fall short of simultaneously satisfying the aforementioned desiderata. Consequently, we devise a dedicated subnet-localization algorithm tailored for efficient fine-tuning that is divided into two stages: \looseness-1

\paragraph{Parameter Importance Calculation} To quantify parameter importance $I(\cdot)$, existing approaches \cite{lecun1989optimal,ma2023llmprunerstructuralpruninglarge} observe the change in loss by assuming $W_k=0$ for the $k$-th parameter. Adopting the second-order Taylor expansion, element-wise importance is estimated as:
\begin{equation}
    \medmath{I=|\frac{\partial \mathcal{L}(\mathcal{D})}{\partial {W_k}}W_k-\frac{1}{2}W_kH_{kk}W_k+o(W_k ^2)|}
    \label{eq:i}
\end{equation}

Here, $H$ stands for the Hessian matrix. However, Eq.\ref{eq:i} is difficult to calculate in real-time. We derive a micro-batch approximation:
\begin{equation}
\medmath{I_i=|\frac{\partial \mathcal{L}(\mathcal{B}_i)}{\partial {W_k}}W_k-\frac{1}{2}(\frac{\sum_j \frac{\partial \mathcal{L}(\mathcal{B}_{ij})}{\partial {W_k}}}{M}W_k)^2 \\+o(W_k ^2)|}
\label{eq:imp}
\end{equation}

Furthermore, estimation with single micro-batch may inject bias by overlooking training dynamics. Sensitivity smoothing and uncertainty quantification \cite{zhang2022platonpruninglargetransformer} are used to handle the problem. At training step $i$, it maintains an exponential moving average (EMA) $\overline{I}_i$ for $I_i$, and uncertainty $\overline{U}_i$ for variation $\Delta I_i=I_i-\overline{I}_i$:

\begingroup
\footnotesize
\begin{align}
\overline{I}_i(W_k) &= \beta_1\overline{I}_{i-1}(W_k) + (1-\beta_1)I_i(W_k) \\
\overline{U}_i(W_k) &= \beta_2\overline{U}_{i-1}(W_k) + (1-\beta_2)|\Delta I_i(W_k)| \\
s(W_k) &= \overline{I}(W_k) \cdot \overline{U}(W_k)
\end{align}
\endgroup

where $\beta_1,\beta_2\in (0,1)$ are the EMA factors. We treat $s(\cdot)$ as an appropriate importance assessment. To obtain the weight-gradient signal without keeping all full tensors in memory, LoSiA uses per-layer updates \cite{lv2024adalomolowmemoryoptimizationadaptive}, executing optimizations during backpropagation without storing gradients.
\paragraph{Core Subnet Localization via Importance Scores} Given a subnet $S$ of the origin network $S_0=(\{i\}_{i=1}^n, \{j\}_{j=1}^m, W)$, define its importance as: 
\begin{equation}
s(S) = \sum_{i\in X_S}\sum_{j\in{Y_S}} s(W_{ij})
\label{eq:sum}
\end{equation}
Our objective is to identify the optimal subnet $S$ that maximizes $s(S)$, while respecting the memory cap $\max\{\frac{|X_S|}{n},\frac{|Y_S|}{m}\}\leq p$, where $p\in (0,1]$ represents the \textit{rank factor}. However, the task is NP-Hard. Exploiting the gradient-magnitude sparsity patterns observed in Section~\ref{sec: grd}, we develop greedy selection algorithms to select the critical input and output neuron set $X_S$ and $Y_S$.

\begin{algorithm}[H]
\renewcommand{\algorithmicrequire}{\textbf{Input:}}
\renewcommand{\algorithmicensure}{\textbf{Output:}}
\caption{Greedy Strategy for Localization}
\label{power}
\begin{algorithmic}[1]
\Require
Importance matrix $s\in\mathbb{R}^{n\times m}$ with $s_{ij}=s(W_{ij})$; 
rank factor $p\in(0,1]$.
\Ensure
$\rho\subseteq\{1,\dots,n\}$ and $\gamma\subseteq\{1,\dots,m\}$ 
denoting the selected input and output neurons.
\Function{Row2Column}{$s\in\mathbb{R}^{n\times m},\ p$}
    \State $\rho\gets\operatorname{Top-K\ Indices}\!\ \bigl(\sum_{j=1}^{m}s_{:,j},\; \lfloor np \rfloor\bigr)$
    \State $\gamma\gets\operatorname{Top-K\ Indices}\!\ \bigl(\sum_{i\in\rho}s_{i,:},\; \lfloor mp \rfloor\bigr)$
    \State \Return $(\rho,\gamma)$
\EndFunction
\end{algorithmic}
\end{algorithm}

\vspace{-5pt}

Algorithm~\ref{power} embodies a row-major greedy policy. First, it locks the $\lfloor np \rfloor$ input neurons with the highest row-wise aggregate importance, then greedily retains the $\lfloor mp \rfloor$ output neurons that maximize the residual mass in those fixed rows. A symmetric column-major variant reverses the order of fixation. The final subnet adopts whichever of the two masks yields the higher score~$s(S)$.

\paragraph{Dimensionality Reduction in Output Layer Fine-Tuning} Although prior work \cite{chen2024longloraefficientfinetuninglongcontext} has established the benefits of fine-tuning the output layer in conjunction with PEFT methods, the approach remains computationally prohibitive for large-vocabulary models (e.g. Gemma-2B). However, empirically, backward propagation through the output layer exhibits gradient sparsity, with only a limited subset of tokens receiving significant updates. Building on this insight, LoSiA easily implements an efficient optimization strategy by constructing a tunable subnet $S=(X_{S_0}, Y_S, W_{X_{S_0},Y_S})$ in the output layer, where $|Y_S| = p_o|Y_{S_0}|$, and $p_o\in (0,1]$ denote the \textit{output dimension reduction factor}.

\subsection{Subnet Optimization and Intergration}
\label{sec:rem}

\begin{figure}[htb]
    \vspace{-5pt}
    \centering
    \includegraphics[width=0.88\linewidth]{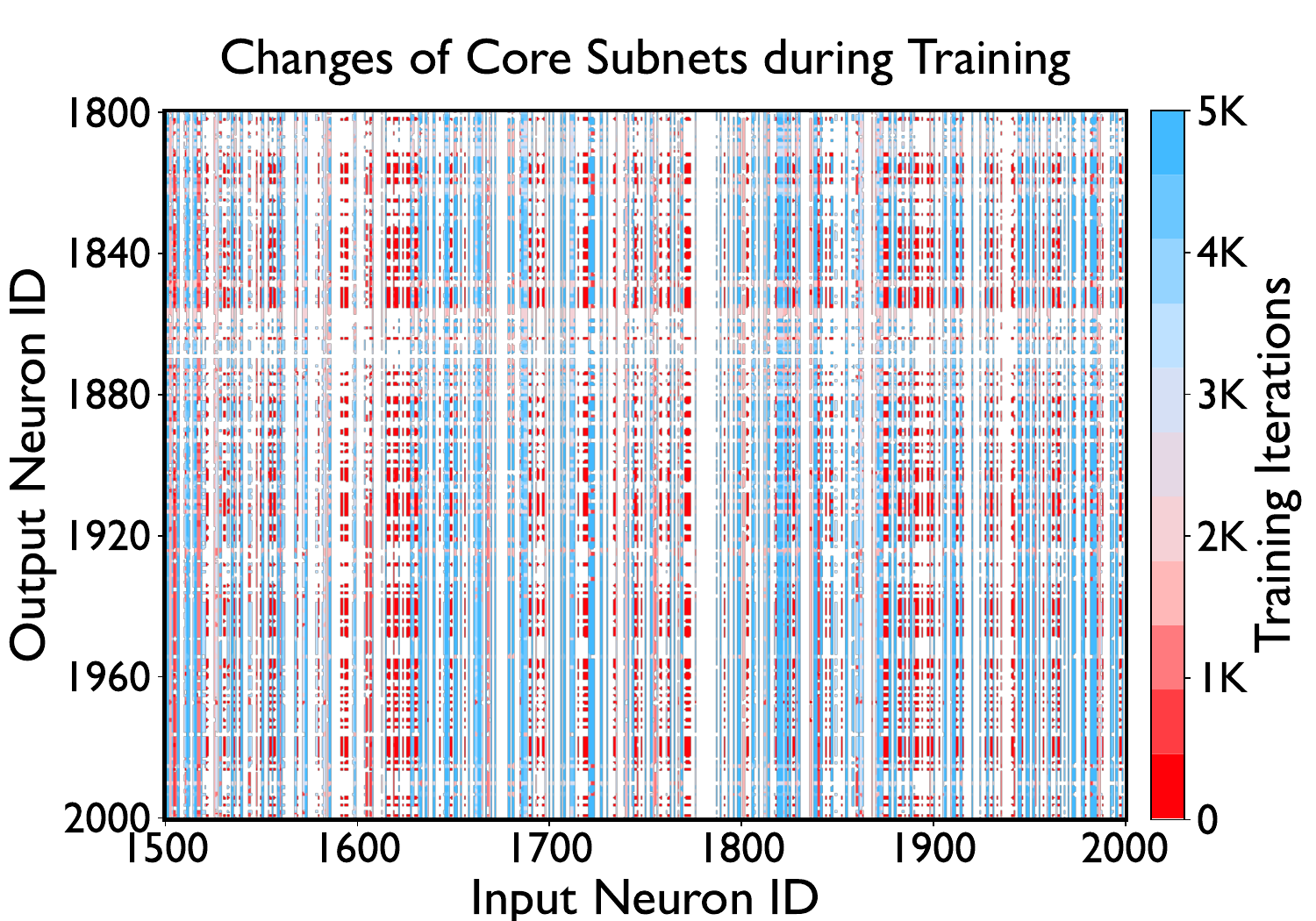}
    \vspace{-5pt}
    \caption{Core Subnet Distribution during Training. The chosen subnets various across training iterations.}
    \label{fig:subnet_dynamics}
\end{figure}

During fine-tuning, the locations of core subnets undergo dynamic shifts, as illustrated in Figure~\ref{fig:subnet_dynamics}.

Although a small subset of neurons is consistently selected, peripheral components exhibit significant temporal variability. Freezing a fixed mask therefore invites under-fitting and over-specialization of the lucky prophase winners. To address the issue, we introduce an asynchronous periodic subnet re-localization mechanism that adapts to the evolving network topology.

Naive periodic learning strategies can induce training instability and loss spikes \cite{lialin2023relorahighranktraininglowrank}. Furthermore, the storage requirements of $\overline{I}(\cdot),\overline{U}(\cdot)$ for every layer simultaneously would lead to a scaling of GPU memory overhead. Therefore, we propose \textit{asynchronous periodic localization} coupled with \textit{rewarmings of learning rate} techniques. Consider a model $f$ with $L$ decoder layers $\{D_l\}_{l=0}^{L-1}$, where each decoder $D_l$ contains $K$ linear layers $\{W_{l,k}\}_{k=1}^K$, with corresponding core subnets $\{S_{l,k}\}_{k=1}^K$. The training timeline is chopped into time slots of length $T$, such that for time slots $[iT,(i+1)T), i=1,2,\ldots$,  we: \looseness-1
\begin{enumerate}
    \item Calculate $\overline{I}(\cdot),\overline{U}(\cdot)$ for layer $D_l$ in time slots $[(kL+l-1)T, (kL+l)T), k\in \mathbb{N}$.
    \item Sequentially reselect $S_l$ by $s(\cdot)$ before step $t = (kL+l)T$, the end of time slots.
\end{enumerate}
Consequently, every core subnet is refreshed exactly once every $\overline{T} = LT$ steps, and, at any moment, only one layer is (i) accumulating importance statistics and (ii) rewarming learning rate. This greatly reduces the extra GPU memory footprint for importance score calculation. 

The rewarming mechanism resets the learning rate to a short warm-up schedule to enhance training stability. Formally, the learning rate at step $t$ is: \looseness-1
\begin{equation}
    \overline{lr}(t) = \begin{cases}
        \frac{t - (kL+l)T}{T} \cdot \text{lr}(t) & \text{if Cond is True} \\
        \text{lr}(t) & \text{otherwise} \\
    \end{cases}
    \\
\end{equation}

\begin{figure}[t]
    \centering
    \includegraphics[width=0.9\linewidth]{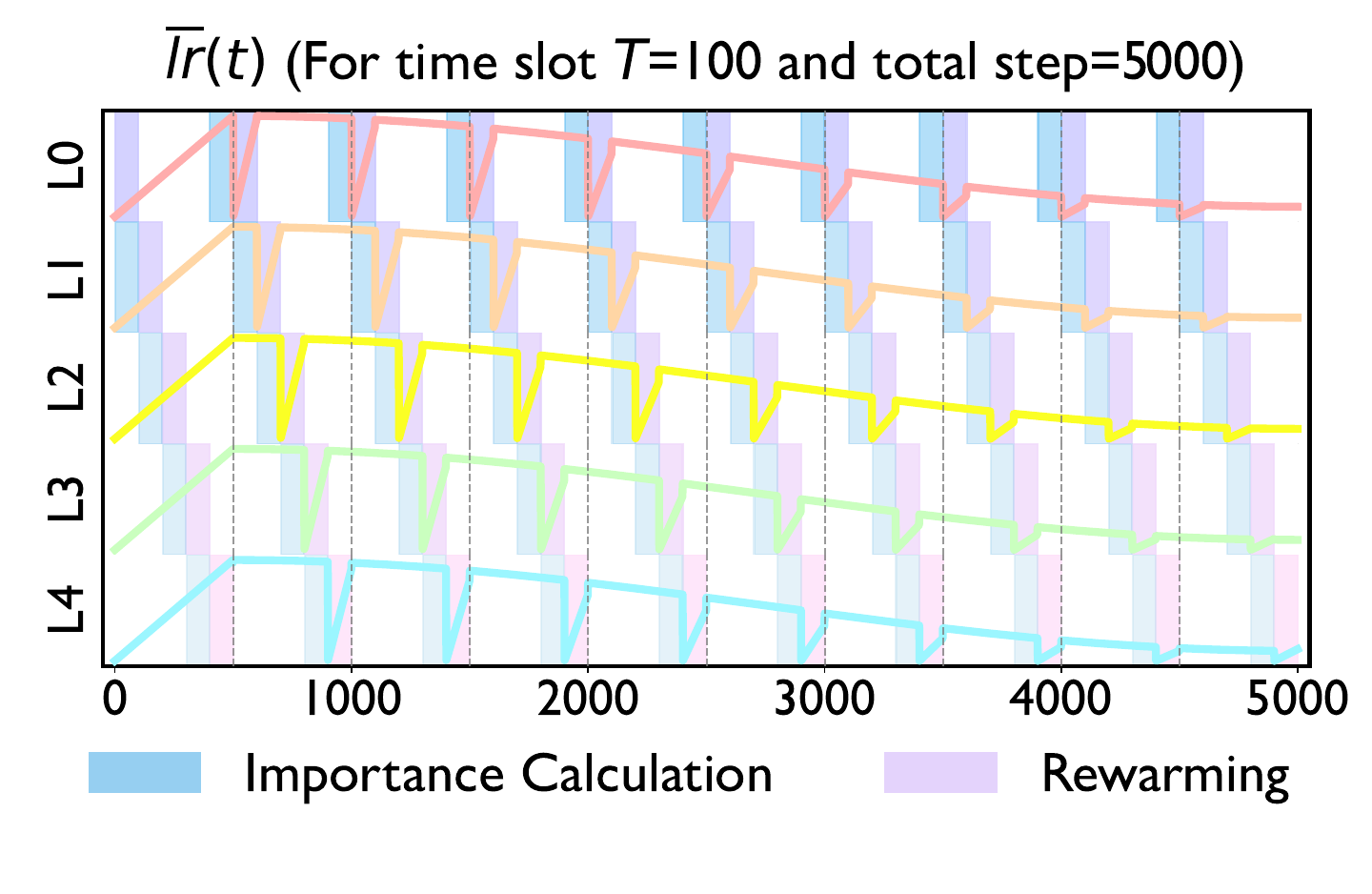}
    \vspace{-10pt}
    \caption{Asynchronous Periodic Subnet Reselection and Learning Rate Rewarming Mechanism (in a 5-layer model for example).\looseness-1} 
    \label{fig:schedule}
    \vspace{-10pt}
\end{figure}

The condition $Cond$ is $t \in [(kL+l)T, (kL+l+1)T) \text{ and } t>T_{w}\ $, where $T_w$ is the warmup duration. This means that rewarmings are triggered only after the initial warmup phase is finished. Figure~\ref{fig:schedule} illustrates the timelines of importance calculation and the rewarming procedure across multiple layers. Importance scores are evaluated, immediately followed by learning rate rewarming, with re-localization sandwiched between them.

\subsubsection{Faster Implementation (LoSiA-Pro)}
Through subnet fine-tuning, LoSiA can further mitigate activation storage and backward latency. The gradient of subnet $S$ can be factorized as:
\begin{equation}
\begin{split}
    \frac{\partial \mathcal{L}}{\partial W_S}&=\frac{\partial \mathcal{L}}{\partial W}[X_S,:][:,Y_S]\\
    &=(x^T[X_S,:])(\frac{\partial\mathcal{L}}{\partial y}[:,Y_S])= \tilde{L}_S\tilde{R}_S
\end{split}
\end{equation} Noticing $\tilde{L}_S\in \mathbb{R}^{np\times bs},\tilde{R}_S\in \mathbb{R}^{bs\times mp}$, the input activation storage is reduced by a factor $p$, while the computational complexity of the gradient calculation is reduced from $O(nmbs)$ to $O(nmbsp^2)$. We named the method LoSiA-Pro, a refined equivalent implementation of LoSiA. It offers a \textbf{27.6\%} latency reduction compared to LoRA, while additionally reducing \textbf{13.4GB} GPU memory consumption compared to LoSiA when training without \textsc{Gradient Check-Pointing}.

\section{Experiments}
We evaluate LoSiA across a wide range of model scales and datasets, conducting rigorous comparisons with common baselines. On both domain-specific and common-sense reasoning tasks, the method demonstrates robust performance with significantly reduced training overheads. The experiments highlight that LoSiA effectively promotes both training efficiency and task proficiency.

\subsection{Experimental Setup}

\paragraph{Datasets}
Models are trained on downstream tasks in the domains of mathematics, coding, and general capabilities. Specifically, training sets are sampled by 50,000 random entries from MetaMathQA, Magicoder, and Alpaca-GPT4, respectively. The GSM8K, MBPP, and MMLU benchmarks are for testing. Additionally, we also compared LoSiA with baseline methods on eight common sense reasoning tasks. More details regarding the datasets can be found in the Appendix.

\begin{table*}[htb]
\centering
\renewcommand{\arraystretch}{1.5}

\caption{Comparison of PEFT Methods Across Models on Domain-Specific Tasks. Accuracy is reported, alongside with memory consumption (GB) and per-token training latency ($\mu s$ / token). The numbers in parentheses indicate the latency of LoSiA-Pro, which is a refined and computationally equivalent implementation of LoSiA.} 
\label{tab:main table}
\resizebox{\textwidth}{!}{
\begin{tabular}{c|c|cc|ccccccc}
\toprule
\multirow{2}{*}{\textbf{Model}} & \multirow{2}{*}{\textbf{Method}} & \multirow{2}{*}{\parbox[c]{1cm}{\centering \textbf{Mem\\(GB)}}} & \multirow{2}{*}{\parbox[c]{2.0cm}{\centering \textbf{Latency\\($\mu s$ / token)}}} & \multicolumn{2}{c}{\textbf{GSM8K}} & \multicolumn{2}{c}{\textbf{MBPP}} & \multicolumn{2}{c}{\textbf{MMLU}}  & \multirow{2}{*}{\textbf{Avg.}}   \\ \cline{5-10}
& & & & 5-shot & 0-shot,CoT & Pass@1 & Pass@10 & 0-shot,PPL & 5-shot,GEN &  \\
 \midrule
 \rowcolor{gray!30} 
\multirow{6}{*}{Gemma 2B}  \cellcolor{gray!0} & FFT & $50.1$ & $142.9$ & 
 $46.4$ & $50.4$ & $33.0$ & $43.4$ & $36.1$ & $37.0$ & $41.05$ \\ 
& LoRA  & $36.1$ & $136.7$ & $35.7$ & $41.1$ & $26.0$ & $36.6$ & $34.9$ & $31.2$ & $34.25$ \\
& PiSSA & $36.1$ & $136.9$ & $38.5$ & $46.5$ & $26.4$ & $39.0$ & $33.8$& $32.6$ & $36.13$ \\
& DoRA  & $37.3$ & $296.6$ & $39.7$ & $43.0$ & $31.4$ & $\mathbf{43.2}$ & $36.2$ & $37.1$ & $38.43$ \\
& GaLore & $37.5$ & $162.4$ & $39.3$ & $44.7$ & $\mathbf{31.6}$ & $42.6$ & $36.6$ & $35.5$ & $38.38$ \\
& \textbf{LoSiA (-Pro)} & $36.9$ & $\mathbf{119.9\ (107.2)}$ & $\mathbf{42.8}$ & $\mathbf{49.7}$ & $30.7$ & $43.0$ & $\mathbf{37.5}$ & $\mathbf{37.4}$ & $\mathbf{40.18}$ \\
\midrule
\rowcolor{gray!30} 
\multirow{6}{*}{LLaMA 2-7B} \cellcolor{gray!0} & FFT & $64.1$ & 359.2 & $46.6$ & $46.9$ & $29.9$ & $40.2$ & $45.2$& $42.5$ & $41.88$  \\
& LoRA  & $23.7$ & $338.0$ & $42.9$ & $46.7$ & $26.0$ & $37.8$ & $42.3$ & $37.3$ & $38.83$ \\
& PiSSA  & $23.7$ & $338.5$ & $43.5$ & $46.2$ & $26.8$ & $36.6$ & $42.7$ & $38.5$ & $39.05$ \\
& DoRA  & $24.2$ & $656.4$ & $\mathbf{45.0}$ & $\mathbf{47.2}$ & $26.0$ & $34.4$ & $44.1$& $36.7$ & $38.90$ \\
& GaLore  & $23.7$ & $437.7$ & $42.2$ & $45.3$ & $28.0$ & $39.0$ & $43.1$& $41.2$ & $39.80$ \\
& \textbf{LoSiA (-Pro)}  & $21.9$ & $\mathbf{290.4\ (244.8)}$ & $44.7$ & $46.7$ & $\mathbf{28.4}$ & $\mathbf{39.4}$ & $\mathbf{45.0}$ & $\mathbf{41.5}$ & $\mathbf{40.95}$\\
\midrule
\rowcolor{gray!30} 
\multirow{4}{*}{LLaMA 2-13B} \cellcolor{gray!0} & FFT-8Bit  & $77.1$ & $640.7$ & $61.2$ & $55.7$ & $35.7$ & $43.2$ & $53.6$ & $56.2$ & $50.93$ \\
& LoRA  &  $36.9$ & $621.1$ & $58.6$ & $\mathbf{56.4}$ & $34.1$ & $44.8$ & $52.6$ & $53.7$ & $50.03$ \\
& PiSSA  & $36.9$ & $622.3$ & $53.4$ & $55.2$ & $34.5$ & $44.8$ & $52.0$ & $48.8$ & $48.11$ \\
& \textbf{LoSiA (-Pro)}  & $36.9$ & $\mathbf{548.5\ (453.4)}$ & $\mathbf{59.0}$ & $54.0$ & $\mathbf{34.9}$ & $\mathbf{48.2}$ & $\mathbf{53.1}$ & $\mathbf{55.7}$ & $\mathbf{50.82}$ \\
\bottomrule
\end{tabular}
}
\vspace{-5pt}
\end{table*}

\paragraph{Implementation Details}

We employ Gemma 2B, LLaMA-2 7B, and LLaMA-2 13B as the backbone models. The effectiveness of LoSiA is evaluated against parameter-efficient fine-tuning (PEFT) baselines, namely LoRA, DoRA, PiSSA, and GaLore. For control of consistency in memory consumption, the rank \(r \) of LoRA, DoRA, and PiSSA is set to $64$. For GaLore, the gradient projection rank \(R\) is set to $512$ with the full projection strategy. In the case of LoSiA, the rank factor $p$ is set to \(\frac{1}{8}\). The learning rate is $6 \times 10^{-5}$ for MetaMathQA and $5 \times 10^{-5}$ for the rest, with time slots $T$ of $100$ for MetaMathQA and $150$ for the rest.

Additionally, both GaLore and LoSiA incorporate the output layer into the fine-tuning process. Dimension reduction factor $p_0$ is set to \(\frac{1}{64}\) for Gemma 2B, \(\frac{1}{8}\) for LLaMA-2 7B, and \(1\) for LLaMA-2 13B in LoSiA. The PEFT modules are applied to all linear layers within the transformer. The training batch size is set to $4$, the warm-up ratio is set to $0.1$ and the model is trained by $3$ epochs.  For training stability, the backbone models are in precision of BF16 and low-rank modules are upcasted to FP32. \footnote{Trained with LLaMA-Factory \cite{zheng2024llamafactory}. Upcasting to FP32 only costs an additional $0.6$GB of memory.} All of the experiments are conducted on single NVIDIA A800 80GB GPU. Further details of the experimental setting (including implementation details on common-sense reasoning tasks) can be found in Appendix \ref{sec:exp}. \looseness-1

\begin{table*}[htb]
\centering
\renewcommand{\arraystretch}{1.3}
\caption{Comparison of PEFT Methods on Commen-Sense Reasoning Tasks, using LLaMA-2 7B as the
 backbone model. Evaluations are PPL-based in lm-evaluation-harness and we report the ACC metric.}
\label{tab:common}
\resizebox{\textwidth}{!}{%
\begin{tabular}{cccccccccccc}
\toprule
\textbf{Method} & \textbf{Mem(GB)} & \textbf{Time(h)} & \textbf{ARC-C} & \textbf{ARC-E} & \textbf{HellaSwag} & \textbf{Winogrande} & \textbf{PIQA} & \textbf{OBQA} & \textbf{SIQA} & \textbf{BoolQ} & \textbf{Avg.} \\ \midrule
LoRA & $19.46$ & $10.0$ & $50.28$ & $79.71$ & $59.86$ & $73.88$ & $79.33$ & $55.00$ & $56.86$ & $\mathbf{88.07}$ & $67.87$ \\
PiSSA & $19.46$ & $10.1$ & $51.19$ & $79.80$ & $62.36$ & $77.74$ & $80.41$ & $56.60$ & $59.88$ & $87.71$ & $69.46$ \\
DoRA & $20.42$ & $25.6$ & $51.71$ & $79.34$ & $59.86$ & $\mathbf{79.24}$ & $79.98$ & $59.60$ & $59.57$ & $88.04$ & $69.67$ \\
GaLore & $\mathbf{18.24}$ & $16.7$ & $48.63$ & $79.97$ & $60.07$ & $76.24$ & $80.09$ & $56.80$ & $56.65$ & $82.60$ & $67.63$ \\
\textbf{LoSiA} & $18.68$ & $\mathbf{9.2}$ & $\mathbf{52.22}$ & $\mathbf{80.26}$ & $\mathbf{65.05}$ & $77.19$ & $\mathbf{81.50}$ & $\mathbf{61.40}$ & $\mathbf{61.05}$ & $84.13$ & $\mathbf{70.35}$ \\
\bottomrule
\end{tabular}
}
\vspace{-10pt}
\end{table*}

\begin{figure}[t]
    \centering
    \includegraphics[width=1.0\linewidth]{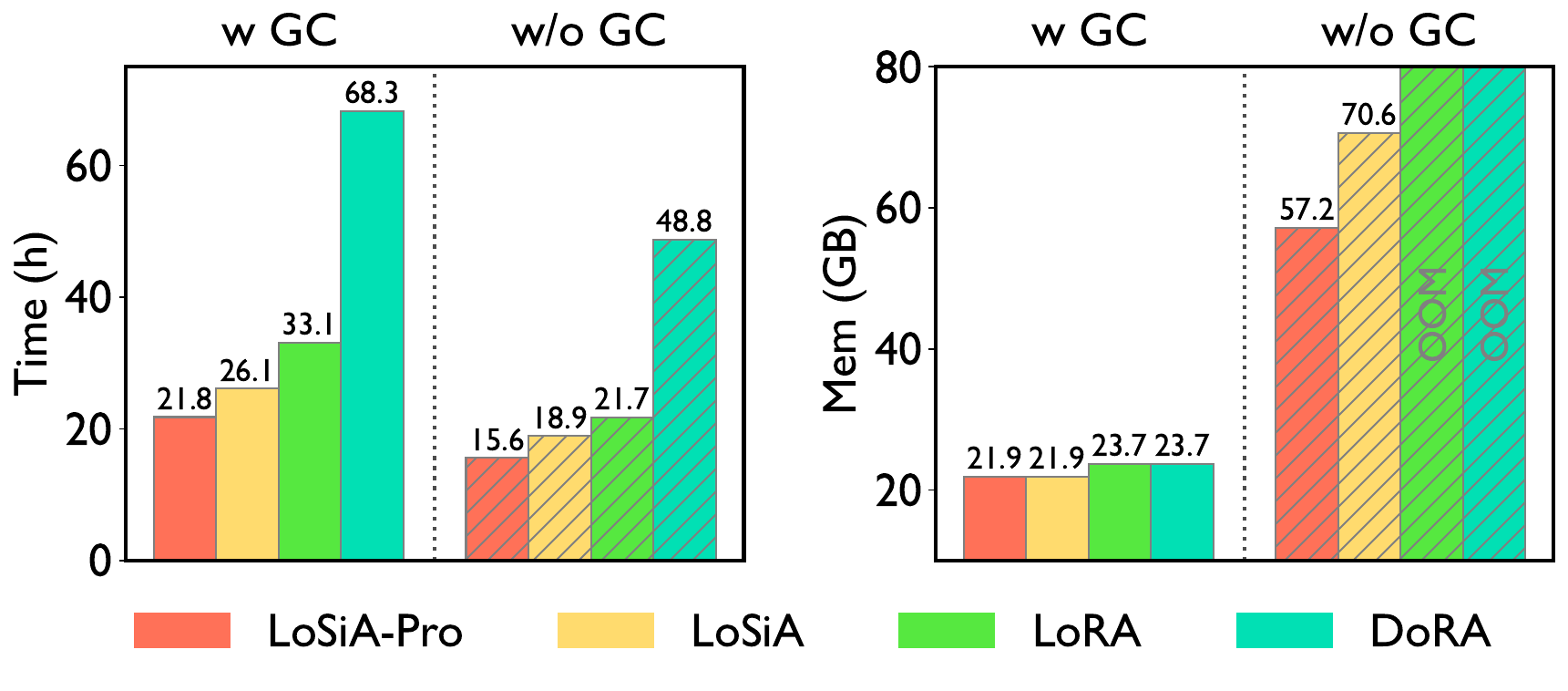}
    \vspace{-20pt}
    \caption{Overheads Comparison of PEFT methods training with and without Gradient Check-Pointing (GC). Taking training arguments in Table \ref{tab:main table} as example.\looseness-1} 
    \label{fig:drawres}
    \vspace{-15pt}
\end{figure}

\subsection{Main Results}

Table \ref{tab:main table} presents the overall performance of LoSiA compared to baseline methods across Gemma-2B, LLaMA2-7B, and LLaMA2-13B models. For GSM8K, we report 0-shot Chain-of-Thought (CoT) and 5-shot accuracy to reveal the model's reasoning capability and few-shot prompting performance. For MBPP, we report the Pass@1 and Pass@10 metrics. For MMLU, we report both 5-shot generation and perplexity-based results. The metrics are intended to measure the quality of generation and knowledge proficiency, respectively. Table \ref{tab:common} shows the 
results on common-sense reasoning tasks, extracting the option with minimum perplexity, and reporting ACC metric following lm-evaluation-harness. The evaluation provides a robust measure of intrinsic knowledge acquisition.\looseness-1

\paragraph{LoSiA effectively reserves knowledge} LoSiA demonstrates superior knowledge retention, as evidenced by perplexity-based evaluations. It outperforms LoRA by 2.48\% on common-sense reasoning tasks and maintains an average 1.93\% improvement on MMLU (0-shot, PPL). Unlike low-rank methods, LoSiA's sparse, high-rank fine-tuning approach enables localized knowledge retention while shifting likelihood toward correct answers.

\paragraph{LoSiA demonstrates superior performance in generalization}  In domain-specific tasks, LoSiA achieves average improvements of 1.75\%, 1.15\%, and 0.79\% compared to the best baseline, respectively. High-rank update methods such as GaLore also exhibit relatively stable performance. The method shows its strength in problem-solving metrics (GSM8K, MBPP Pass@1, and MMLU 5-shot), suggesting that LoSiA provides strong generalization capabilities by applying learned knowledge to address various problems.  Notably, while performing comparable to Full-Parameter Fine-Tuning (FFT) with only 0.64\% of degradation in average, LoSiA significantly reduces computational resources, highlighting its practical efficiency.

\paragraph{LoSiA and LoSiA-Pro greatly improve training efficiency}

Figure \ref{fig:drawres} compares the training overheads of various PEFT methods. Contrast to baselines such as DoRA that incur significant additional FLOPs, LoSiA shows superior efficiency in both training latency and memory usage. By eliminating extra matrix multiplication operations, LoSiA achieves faster training speeds. Its refined implement, LoSiA-Pro, further compresses activation storage by at least \textbf{22.8GB} (w/o GC) and raise training throughput to \textbf{1.38x} (w GC) compared to LoRA by saving and computing on partial activations. A detailed training latency and GPU memory measurement is in Appendix \ref{sec:speed}.

\subsection{Ablation Study}
This section assesses the functionality of sensitivity importance-aware localization, asynchronous mechanism, re-warmings, and re-localizations, alongside with robustness analysis of LoSiA. We present comprehensive ablation studies in Table \ref{tab:ablation} and training dynamics in Figure \ref{fig:ablation loss curve}. Additional robustness tests for rank factor selection are provided in the Appendix.

\begin{figure}[t]
    \centering
    \includegraphics[width=1.0\linewidth]{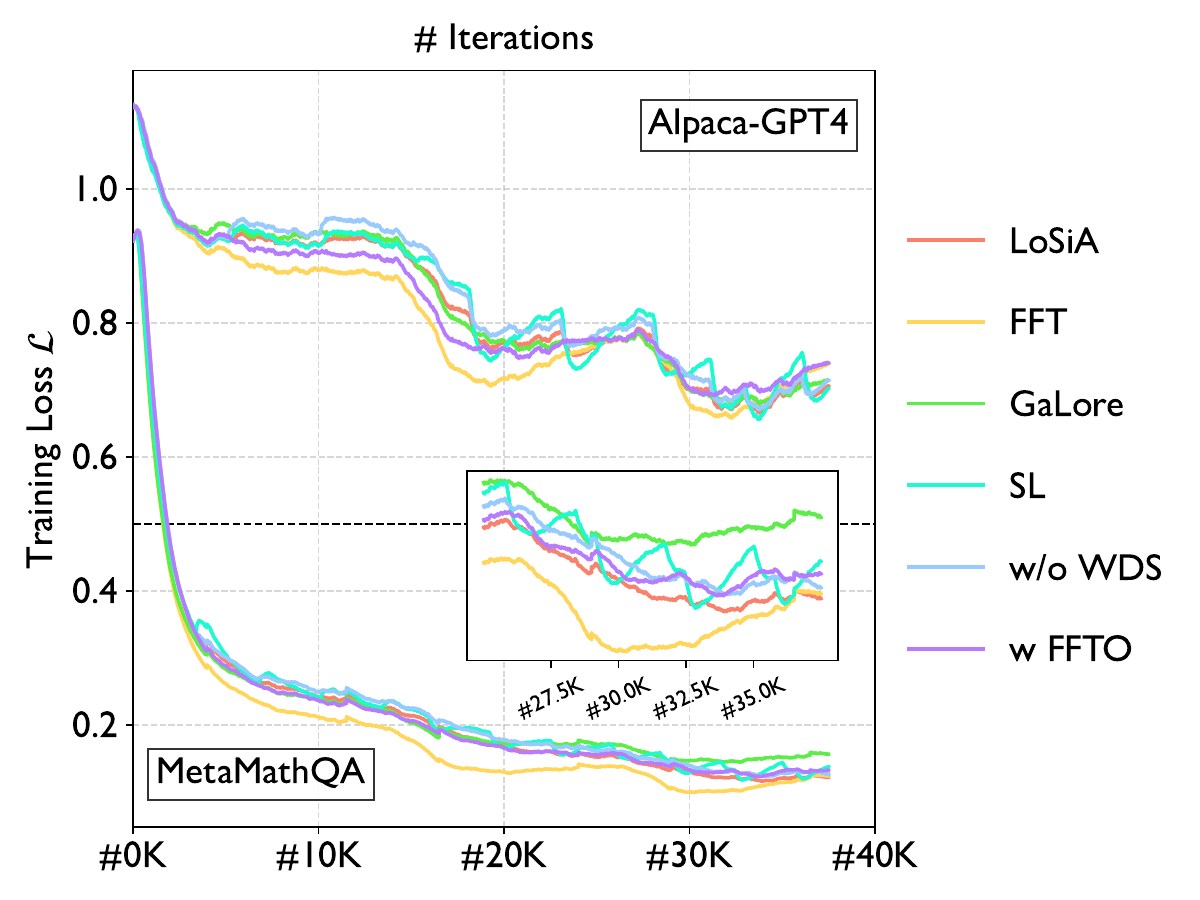}
    \vspace{-10pt}
    \caption{Loss Curves of Baselines and LoSiA Variants, training on MetaMathQA and Alpaca-GPT4.}
    \label{fig:ablation loss curve}
    \vspace{-15pt}
\end{figure}

\begin{table}[htb]
\centering
\renewcommand{\arraystretch}{1.1}
\caption{Ablation Study of LoSiA on GSM8K and MMLU benchmark, using LLaMA-2 7B as
 backbone. }
\label{tab:ablation}
\resizebox{0.9\columnwidth}{!}{%
\begin{tabular}{cccc}
\toprule
\textbf{Model} & \textbf{GSM8K} & \textbf{MMLU} & \textbf{Avg.} \\
\midrule \addlinespace[0.4em] 
\textbf{Vanilla LoSiA} & $\mathbf{44.66}$ & $\mathbf{44.95}$ & $\mathbf{44.81}$ \\ \addlinespace[0.3em] \hline \addlinespace[0.3em]
\makecell{Synchronous \\ Localization (\textit{SL})} & $42.76$ & $44.13$ & $43.45$ \\ \addlinespace[0.3em]  \hline\addlinespace[0.3em] 
\makecell{Gradient-based \\ Localization (\textit{GL)}} & $43.00$ & $44.88$ & $43.94$ \\ \addlinespace[0.3em] \hline \addlinespace[0.3em] 
\makecell{w/o Warm-up \\during Selection (\textit{WDS})} & $38.06$ & $44.21$ & $41.14$ \\ \addlinespace[0.3em]  \hline \addlinespace[0.3em] 
\makecell{w FFT lm\_head \\(\textit{FFTO})} & $43.96$ & $44.32$ & $44.14$ \\ \hline
\addlinespace[0.3em]
\makecell{w/o Re-localization \\(\textit{ReLO})} & $42.76$ & $43.81$ & $43.29$ \\
\bottomrule
\end{tabular}
}
\vspace{-5pt}
\end{table}

\paragraph{Asynchronous re-localization yields more stable training} When each layer fine-tunes with fixed core subnets as \textit{ReLO} represents, it results in persistent under-fitting and marked drop in test accuracy, confirming that key parameters shift frequently during training, necessitating dynamic and periodic localization of the selected core sub-network to adapt in real time. Variant \textit{SL} refers to using a synchronous layer-wise localization strategy. However, it causes loss fluctuation, destabilizes later training, and degenerates the model performance by 1.36\% on average, while asynchronous updates produce more stable loss curves.

\paragraph{Sensitivity-based importance versus gradient-based importance} Variant \textit{GL} adopts absolute gradients as the importance score. On MMLU, its performance remains comparable to LoSiA but is biased towards humanities tasks (see Table \ref{tab:glsl}), while its accuracy on GSM8K drops by 1.66\%. Sensitivity-based scores, which aggregate multi-sample information, are more effective in capturing general patterns in linear layers compared to biased gradients. However, the gradient-based variant exhibits promising results. In practice, the storage of $\overline{I}(\cdot),\overline{U}(\cdot)$ (about 1GB of memory occupation on LLaMA-2 7B) can be eliminated using gradient-based importance if needed. Further discussion is provided in Appendix \ref{sec:score}.

\paragraph{Effect of rewarming and full fine-tuning the output layer} The variant \textit{w/o WDS}, which omits rewarm-ups, introduces instability of the loss, leads to under-fitting and ultimately impairs final performance. \textit{w FFTO} fully fine-tunes the output layer, shows a performance comparable to LoSiA with additional trainable parameters. It highlights the effectiveness of extracting tunable subnets on the output layer in LoSiA. In permissible GPU memory constraints, fully training the output layer shows promising performance and is also recommended.

\begin{table}[htb]
\centering
\renewcommand{\arraystretch}{1.1}
\caption{Robustness of Time Slot $T$ Across a Series of Data Scales. Trained with MetaMathQA and evaluated by GSM8K on LLaMA-2 7B. }
\vspace{-2pt}
\label{tab:datascale}
\resizebox{0.65\columnwidth}{!}{%
\begin{tabular}{ccccccc}
\toprule
\textbf{Method} & \textbf{@30K} & \textbf{@50K} & \textbf{@70K} \\ 
\midrule
\textit{LoRA} & $41.39$ & $42.86$ & $44.58$ \\
\midrule
$T$ & \multicolumn{3}{c}{\textit{LoSiA}} \\
\midrule
$25$ & $\mathbf{42.99}$ & $\mathbf{43.37}$ & $42.07$ \\
$50$ & $\mathbf{42.91}$ & $42.46$ & $42.15$ \\
$75$ & $41.09$ & $\mathbf{44.05}$ & $\mathbf{47.46}$ \\
$100$ & $40.49$ & $\mathbf{44.66}$  & $\mathbf{46.17}$ \\
$125$ & $39.88$ & $42.23$ & $\mathbf{45.19}$ \\
$150$ & $39.12$ & $40.41$ & $42.84$ \\
\bottomrule
\end{tabular}
}
\vspace{-10pt}
\end{table}

\paragraph{Robustness across varying data scales and time slot lengths} Table~\ref{tab:datascale} benchmarks LoSiA against LoRA as the training corpus grows. Across every scale, LoSiA consistently outperforms LoRA, demonstrating stability and robustness. Furthermore, the optimal time slot $T$ is positively correlated with the size of the training set, while LoSiA shows transcendent performance within a reasonable range of $T$. \looseness-1

\subsection{Analysis}
\begin{figure}[htb]
    \centering
    \includegraphics[width=1.0\linewidth]{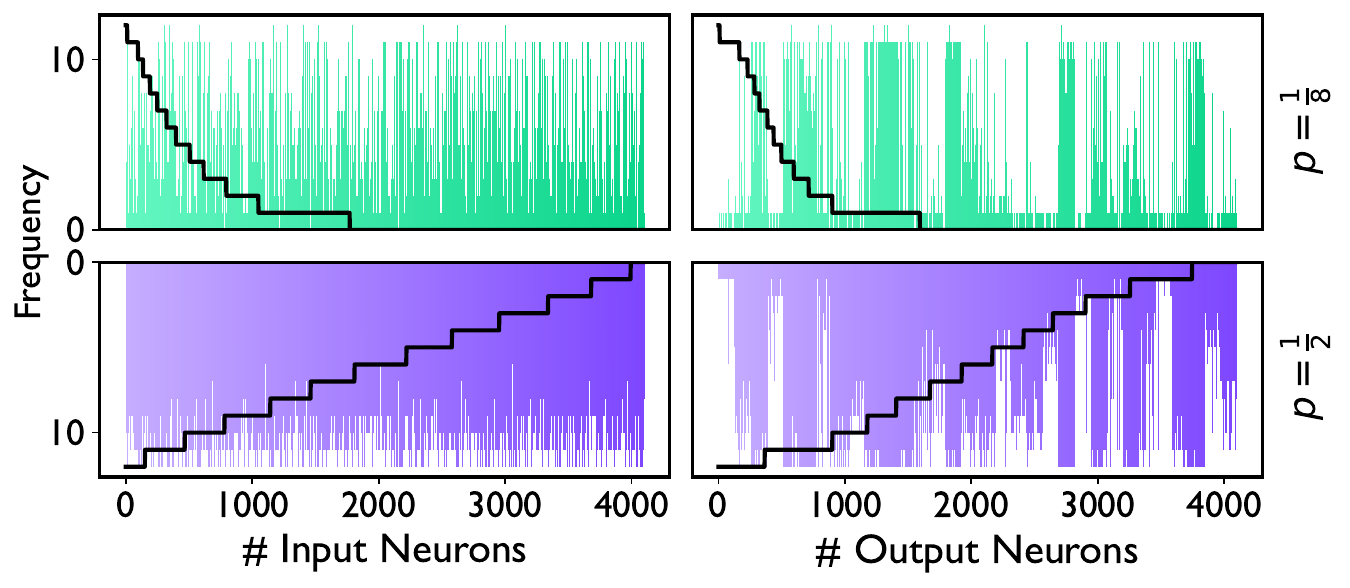}
    \vspace{-20pt}
    \caption{Selected Frequency Distributions of Neurons in Core Subnets. Sorted frequencies are ploted in black.}
    \label{fig:sec}
    \vspace{-10pt}
\end{figure}

\paragraph{Selection Distribution} Figure~\ref{fig:sec} visualizes how often each neuron is chosen in core subnets during training. The frequently selected neurons remain similar under different rank factors $p$. The smaller $p$ merely sharpens the histogram: mass is pushed into the most important weights, so more radical compression does not discard salient parameters. LoSiA simultaneously adjusts marginal parameters to enhance generalization capability, yielding better generalization whenever the budget is tight.

\begin{figure}[htb]
    \centering
    \includegraphics[width=1.0\linewidth]{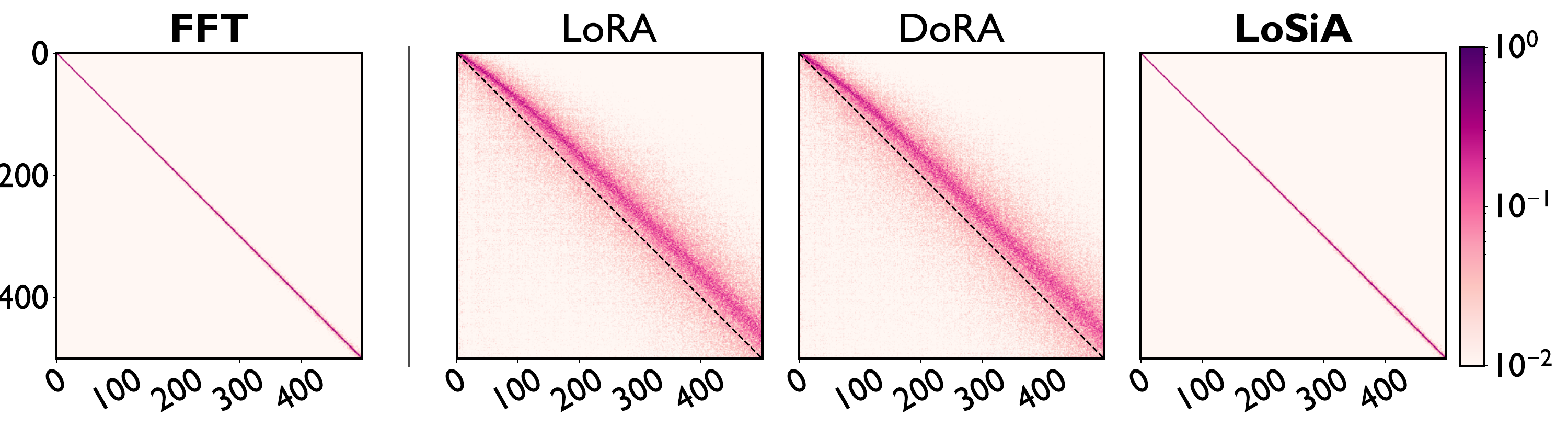}
    \vspace{-10pt}
    \caption{Similarities of Top-500 Largest Singular Vectors between Pre- and Post-Fine-Tuning Weights.}
    \label{fig:sim}
    \vspace{-10pt}
\end{figure}

\paragraph{Reduce Intruder Dimensions} Low-rank fine-tuning methods often introduce intruder dimensions \cite{shuttleworth2024lora}, resulting in spectral discrepancies between the fine-tuned and the pre-trained weights. This diminishes the adaptability of LoRA in sequential learning. Figure \ref{fig:sim} illustrates the cosine similarity between the $\text{Top-}500$ singular vectors of the trained matrices and those of the original weights. Both LoRA and DoRA exhibit dimensional shifts due to their low-rank structures, whereas LoSiA demonstrates higher similarity and dimensional stability comparable to FFT.

To evaluate LoSiA's efficacy in continual learning, we perform sequential fine-tuning on Hellaswag, PiQA, BoolQ, SiQA, and Winogrande datasets on LLaMA-2 7B. We employ Average Performance (AP) \cite{DBLP:journals/corr/abs-1801-10112}, Forward Transfer (FWT) \cite{DBLP:journals/corr/Lopez-PazR17}, and Backward Transfer (BWT) \cite{ke2023continuallearningnaturallanguage} metrics to assess overall performance, knowledge transfer ability from previous tasks to current task, and level of forgetting, respectively. Details of the experiments are provided in Appendix \ref{sec:cl}.

Suppose that the model learns sequentially on $N$ tasks. Let $P_{i,j}$ denote the accuracy on task $j$ after training on task $i$. Following \citet{zhao2024saptsharedattentionframework, feng2025recurrent}, we formulate the metrics (AP, FWT and BWT) as bellow:

\textbf{Average Performance}: The metric reflects overall task performance after continued learning, which is, $AP=\frac{1}{N}\sum_{i=1}^N P_{N,i}$

\textbf{Forward Transfer}: The metric measures the transferability of learned knowledge from previous tasks to a new task. $\text{FWT}=\frac{1}{N}\sum_{i=1}^N(P_{i,i}-P_{0,i})$, where $P_{0,i}$ is the performance of individually training task $i$.

\textbf{Backward Transfer}: The metric evaluates the impact of learning later tasks on the model’s performance on an earlier task, that is, $\text{BWT}=\frac{1}{N-1}\sum_{i=1}^{N-1}(P_{N,i}-P_{i,i})$. 

\begin{table}[htb]
\centering
\renewcommand{\arraystretch}{1.0}
\caption{Results of Continue Learning with Sequential PEFTs on Five Commen-Sense Reasoning Tasks.\looseness-1}
\label{tab: cl}
\resizebox{0.8\columnwidth}{!}{%
\begin{tabular}{cccc}
\toprule
\textbf{Method} & \textbf{AP($\uparrow$)} & \textbf{FWT($\uparrow$)} & \textbf{BWT($\uparrow$)} \\ 
\midrule
Seq-LoRA & $66.62$ & $\mathbf{1.46}$ & $-8.04$ \\
Seq-LoSiA & $\mathbf{70.48}$ & $-0.20$ & $\mathbf{-3.54}$ \\
\bottomrule
\end{tabular}
}
\end{table}

The results in Table \ref{tab: cl} demonstrate that LoSiA outperforms LoRA in mitigating forgetting with 4.5\% in BWT and achieves a 3.86\% improvement in average performance of sequential fine-tuning.  This aligns with our hypothesis that LoSiA exhibits stronger robustness in continue learning, indicating that our method can adapt to more diverse application scenarios than existing baselines.

\section{Conclusion}

\looseness=-1
We present LoSiA, a novel PEFT framework that dynamically identifies and optimizes core sub-networks. Through sensitivity-based localization, asynchronous re-selection, and efficient high-rank adaptation, LoSiA achieves high throughput and low activation overhead. Extensive experiments show that LoSiA outperforms baselines on domain-specific and common-sense reasoning tasks while reducing forgetting.
We hope that our work will inspire future research to further explore intrinsic substructures in supervised fine-tuning.

\section{Limitation}
The innovative design of locating and optimizing sub-networks enables LoSiA to demonstrate outstanding advantages in terms of efficiency and performance. This work preliminarily validates the effectiveness of fine-tuning focused on substructures, yet there remains considerable room for further exploration and improvement. The effectiveness in scenarios such as multi-tasking, vision, and format alignment remains unclear. As for the method, the subnet localization in LoSiA is relatively rigid, and may still fail to precisely capture all critical neuron connections. More flexible and accurate approaches for the location of substructures, such as dynamically adjusting the rank factor for various layers, could further enhance performance.

Furthermore, while LoSiA can be conveniently integrated with other training platforms, additional efforts are required to improve its usability in real-world production scenarios. Currently, our work aims to provide individuals and small enterprises with a highly efficient single-GPU fine-tuning method, but the workflow could be further extended to multi-GPU environments. Moreover, to accommodate diverse datasets and practical deployment conditions, automated time slot selection mechanisms warrant further investigation.

\section*{Acknowledgements}
This work is supported by the National Natural Science Foundation of China (No. 52539001).  It's partially supported by CHN Energy Dadu River Big Data Services Co., Ltd. It’s also supported by Student Research Training (SRT) project of Tsinghua University. We also thank anonymous reviewers for their valuable feedback. 
\bibliography{custom}

\newpage
\appendix
\section{Appendix}
\label{sec:appendix}

\subsection{Derivations and Proofs}
\subsubsection{Proof for Formula \ref{eq:mse}}
\label{sec: grdprf}
On a batch $\mathcal{B}$ composed of $M$ samples, the MSE loss between full fine-tuning (which produces model $f$) and training on parameter set $P$ (which produces model $f'$) is given by:
\begin{align}
    \mathcal{L}_{MSE} & = \frac{\|y-y'\|^2_F}{M} = \frac{\|Wx-W'x\|^2_F}{M} \\
    & \leq \frac{\|W-W'\|^2_F\|x\|^2_F}{M}
\end{align}
\paragraph{SGD} In SGD optimizer, supposing the learning rate is $\eta$, the difference in fine-tuned parameter is:
\begin{align}
    W-W' = - \eta1_{(i,j)\not \in P}\cdot \nabla_{W_0} \mathcal{L} (\mathcal{B})
\end{align}
It derives an upper bound for the MSE Loss:
\begin{align}
    \mathcal{L}_{MSE} & \leq \eta ^2\frac{\|1_{(i,j)\not \in P}\cdot \nabla_{W_0} \mathcal{L} (\mathcal{B})\|^2_F\|x\|^2_F}{M}
\end{align}

The result suggests that maximizing the sum of $\nabla_{W_0}\mathcal{L} (\mathcal{B})_{ij}$ where $(i,j)\in P$ ideally tightens the approximate error of training on parameter subset.

\paragraph{AdamW} In AdamW optimizer, at training step $t$, the first-order momentum $M_t$ and second-order momentum $V_t$ are calculated by:
\begin{align}
    G_t = & \nabla_{W} \mathcal{L} (\mathcal{B}_t) \\
    M_t = & \beta_1M_{t-1}+(1-\beta_1) G_t \\
    V_t = & \beta_2V_{t-1}+(1-\beta_2) G_t^2 \\
    \tilde{G_t} = & \frac{M_t}{\sqrt{V_t+\epsilon}}
\end{align}

Similarly, since $W-W'=-\eta1_{(i,j)\not \in P}\cdot \tilde{G_t}$, we analyze the relationship between $\tilde{G_t}$ and the original gradient $G_t$ by element:
\begin{equation}
\begin{split}
    \frac{\partial (\tilde{G_t})^2}{\partial G_t} = & 2M_t[\frac{(1-\beta_1)V_t}{V_t^2} \\
    & - \frac{(1-\beta_2)G_tM_t}{V_t^2}]
\end{split}
\end{equation}

Suppose $M_t>0$, when $G_t < \frac{(1-\beta_1)V_t}{(1-\beta_2)M_t}$, $\frac{\partial (\tilde{G_t})^2}{\partial G_t} > 0$. In practice, typical settings are $\beta_1=0.9, \beta_2=0.999$. Therefore, when $G_t < 10^2\frac{V_t}{M_t}$, $\tilde{G}_t$ increases with $G_t$, effectively covering a broad range of non-stationary optimization scenarios.

\subsubsection{Proof for Formula \ref{eq:imp}}
The foundational work was established by \citet{lecun1989optimal} and \citet{DBLP:journals/corr/KirkpatrickPRVD16}. However, to establish real-time importance calculation during training, approximations are necessary and are derived below. Element-wise importance score $I(\cdot)$ is formulated as:
\begin{equation}
\begin{split}
I(W_k)=&|\Delta\mathcal{L}(\mathcal{D})| = |\mathcal{L}(\mathcal{D})-\mathcal{L}_{W_k=0}(\mathcal{D})| \\
 =& |\frac{\partial \mathcal{L}^T(\mathcal{D})}{\partial {W_k}}W_k-\frac{1}{2}W_kH_{kk}W_k
\\&+o(W_k^2)|
\end{split}
\end{equation}
where $H$ denotes the Hessian matrix, which is computationally intensive. Therefore, the fisher information matrix $F$ is used instead to obtain diagonal elements of the Hessian matrix: 
\begin{equation}
\begin{split}
    F_{kk}=&-H_{kk}=-\mathbb{E}_{p(\theta|\mathcal{D})}[\frac{\partial^2\mathcal{L}(\theta,D)}{\partial^2 \theta_k}|_{\theta=\theta^*}] \\
    \approx & -\mathbb{E}_{(x,y)\sim \mathcal{D}}[(\frac{\partial \mathcal{L}(\theta,x,y)}{\partial\theta_k}|_{\theta=\theta^*})^2]
\end{split}
\end{equation}
Approximating mathematical expectations on dataset $\mathcal{D}$ using the Monte Carlo method derives:
\begin{equation}
\begin{split}
    I(W_k)
 =& |\frac{\partial \mathcal{L}(\mathcal{D})}{\partial {W_k}}W_k+o(W_k^2) \\&-\sum_{(x,y)\in \mathcal{D}}\frac{1}{2|\mathcal{D}|}(\frac{\partial \mathcal{L}(x,y)}{\partial {W_k}})^2W_k^2|
\end{split}
\end{equation}

During training, the dataset $\mathcal{D}$ is processed in batches $\mathcal{B}_i$, and the batch gradient is calculated as $\nabla_W\mathcal{L}(\mathcal{B}_i)=\frac{1}{M}\sum_{j=1}^M \nabla_W\mathcal{L}(\mathcal{B}_{ij})$. To avoid calculating the gradients separately for each sample in the batch, we approximate $\sum_{(x,y)\in \mathcal{B}_i}\frac{1}{M}(\frac{\partial \mathcal{L}(x,y)}{\partial {W_k}})^2$ to the term of $(\frac{\sum_{(x,y)\in \mathcal{B}_i}\frac{\partial \mathcal{L}(x,y)}{\partial {W_k}}}{M})^2$. To analyze errors, assume $g=\frac{\partial \mathcal{L}(x,y)}{\partial {W_k}}\sim G$, we have:
\begin{equation}
\begin{split}
    \Delta =& |\frac{1}{M}\sum_{j=1}^M g_j^2-(\frac{\sum_{i=1}^M{g_j}}{M})^2| \\=& \frac{1}{M}\sum_{i=1}^M g_j^2-(\frac{\sum_{j=1}^M{g_j}}{M})^2 \\ \leq& \frac{(\max g_j-\min g_j)^2}{4} = O(g^2)
\end{split}
\end{equation}
The approximation errors are bounded. We take the following for importance estimation:

\begin{equation}
    I_i=|\frac{\partial \mathcal{L}(\mathcal{B}_i)}{\partial {W_k}}W_k-\frac{1}{2}(\frac{\sum_j \frac{\partial \mathcal{L}(\mathcal{B}_{ij})}{\partial {W_k}}}{M}W_k)^2 \\+o(W_k ^2)|
\end{equation}
\subsubsection{Maximizing Formula \ref{eq:sum} is NP-Hard}
\paragraph{Task} \textit{Given an arbitrary non negative matrix $A^{n\times m}$ and cardinal budget requirements $\tilde{n},\tilde{m}$. Select $\tilde{n}$ rows $X_S$ and $\tilde{m}$ columns $Y_S$ to maximize the sum $\sum_{i\in X_S}\sum_{j\in Y_S} A_{ij}$.}

\paragraph{Lemma} (The Maximum Clique Problem is NP-Complete) 
Given an undirected graph \( G = (V, E) \), where:
\begin{itemize}
    \item \( V \) is a set of vertices,
    \item \( E \subseteq V \times V \) is a set of edges,
\end{itemize}
a clique \( C \subseteq V \) is a subset of vertices such that every two distinct vertices in \( C \) are adjacent, i.e.,
\[
\forall u, v \in C \ ,u \neq v \Rightarrow (u, v) \in E.
\]
The Maximum Clique Problem (MCP) seeks a clique of maximum cardinality in \( G \). The problem is \textbf{NP-complete}, meaning:
\begin{itemize}
    \item It is \textbf{NP}: a candidate solution can be verified in polynomial time, and
    \item It is \textbf{NP-Hard}: any problem in NP can be reduced to it in polynomial time.
\end{itemize}

\paragraph{Proof} Construct a special form of $A^{n\times n}$ as the adjacent matrix of graph $G$ with larger values on the diagonal maximum, that is:
\[
A_{uv} =
\begin{cases}
1 & \text{if } (u, v) \in E, \\
n^2 + 1 & \text{if } u=v\\
0 & \text{otherwise}
\end{cases}
\]
Then, the MCP problem can be reduced to the task in polynomial time following the algorithm below:

\begin{itemize}
    \item Enumerate $k$ in descending order $n,n-1\ldots1$
    \item Solve the task with $\tilde{n}=\tilde{m}=k$
    \item If the optimal solution equals to $(n^2+k)k$, then there exist a clique $C=X_S$ of size $k$, terminate.
\end{itemize}

Therefore, an NP-Complete problem can be reduced to the task in polynomial time, which yields the conclusion that the task is NP-Hard.

\subsection{Further Observations}
\subsubsection{Gradient Magnitude Distribution}
\label{sec: grdobs}
\begin{figure*}[htbp]
\centering
    \begin{subfigure}{\textwidth}
      \centering
      \includegraphics[width=\linewidth]{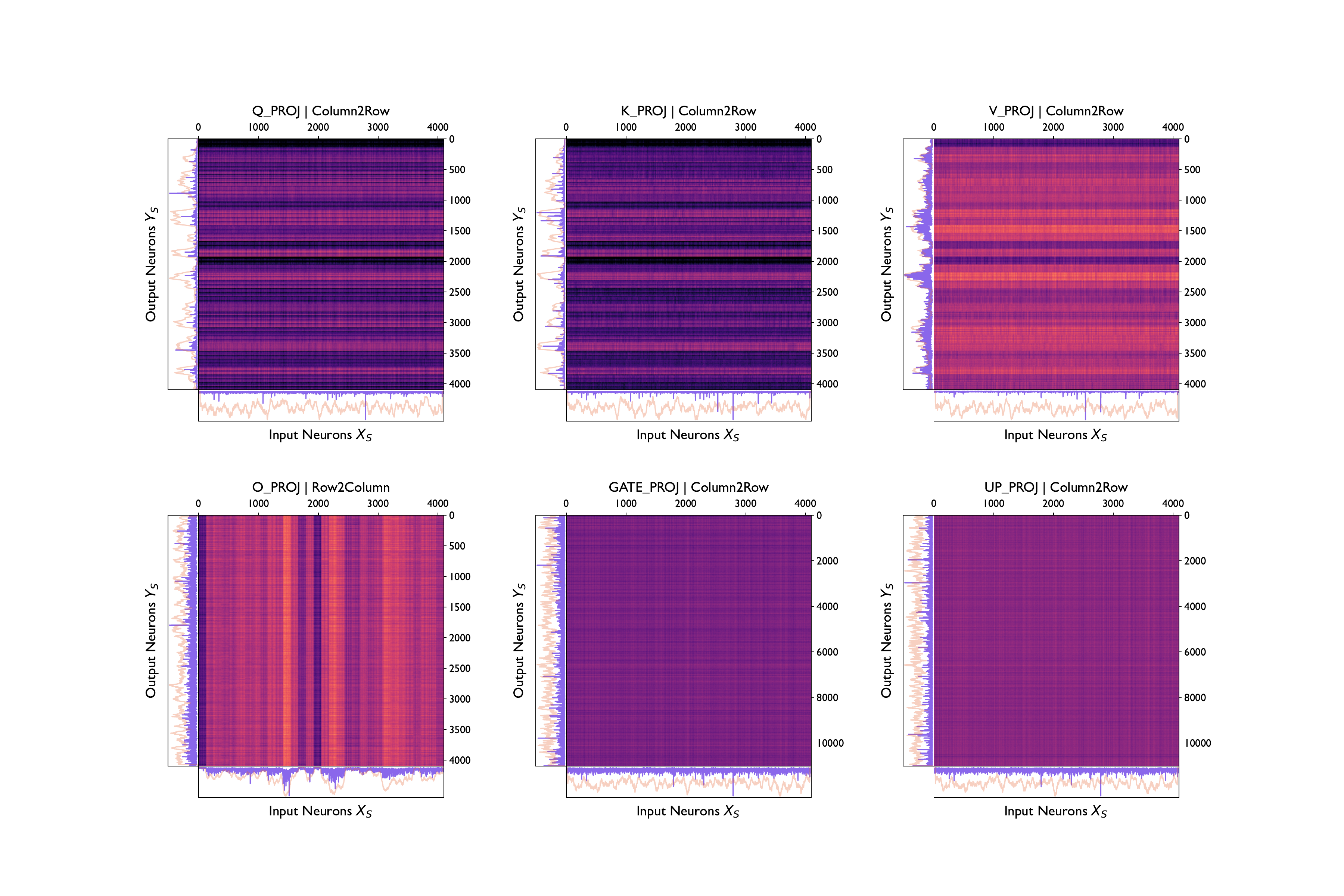}
      \caption{Decoder Layer $15$}
    \end{subfigure}
    \vspace{1em} 
    \begin{subfigure}{\textwidth}
      \centering
      \includegraphics[width=\linewidth]{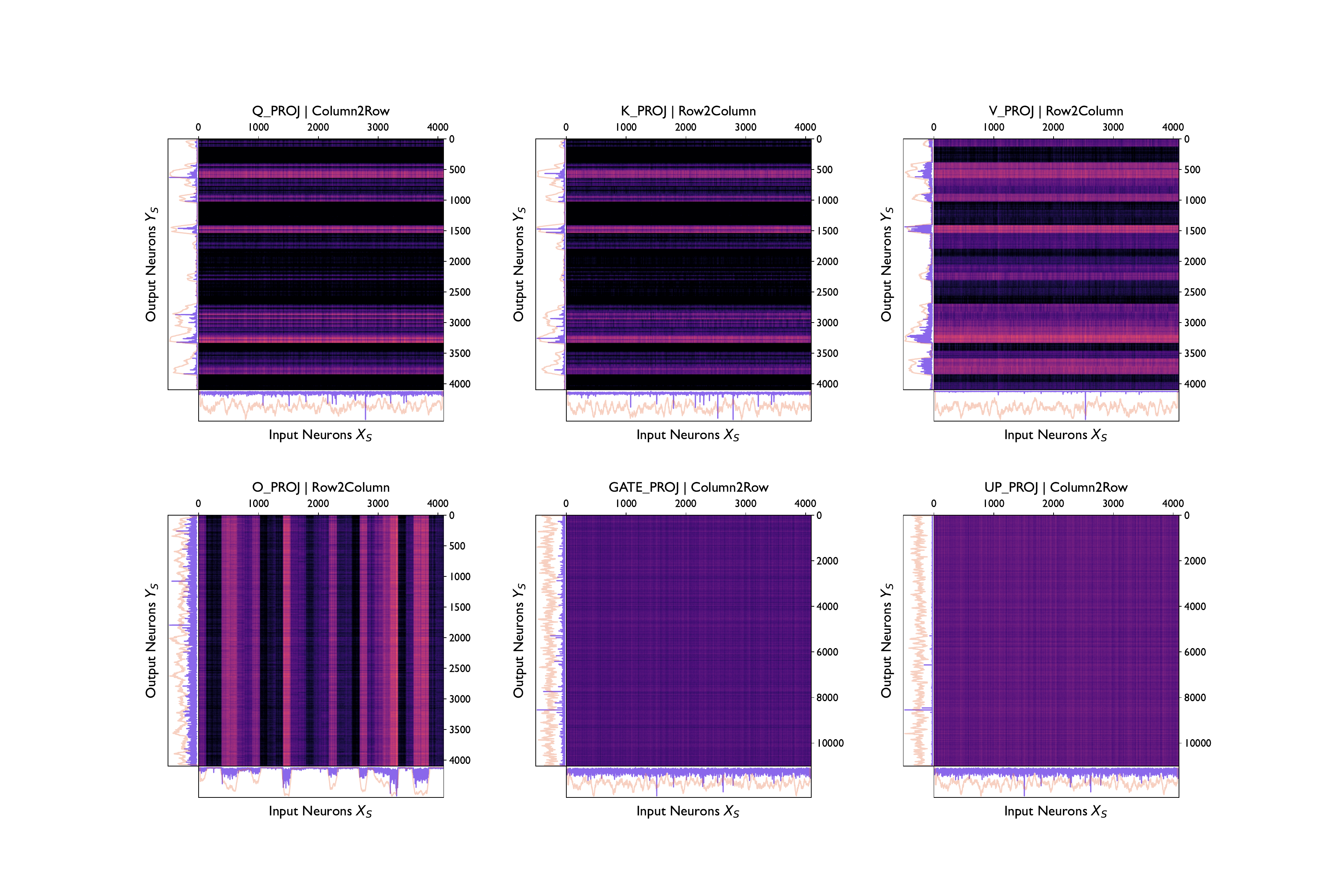}
      \caption{Decoder Layer $25$}
    \end{subfigure}
\caption{Gradient Magnitude Distribution on LLaMA-2 7B for Different Decoder Layers and Modules. Purple curve: row/column gradient sums. Orange curve:
smoothed neuron selecting frequency. Best selection strategy for each layers (Row2Column/Column2Row) are record in the title of subplots.}
\label{fig:wide}
\end{figure*}

To investigate the universality of the sparse sub-network structure for large gradients, we analyze  gradient magnitude distributions across different layers, as shown in Figure \ref{fig:wide}. Both Gradients of the Self-Attention and MLP modules exhibit the consistent structure of core subnets. We also quantify this observation on different layers and each submodule in Table \ref{tab:grad_all}. Core-subnet localization markedly outperforms random selection and closely approaches the ideal Top-K baseline, corroborating the validity of sparse-gradient patterns. Note, however, that the ideal Top-K set is irregular and therefore entails nontrivial runtime overhead.

\begin{table}[htbp]
  \centering
  \small
  \caption{Sum of Absolute Gradient Values ($\times 10^3$) for Selection Patterns. "Subnet" stands for the core subnet localization algorithm, while Top-K is an ideal selection.}
  \label{tab:grad_all}
  \setlength{\tabcolsep}{4pt}
  \resizebox{1.0\columnwidth}{!}{%
  \begin{tabular}{@{}cccccc@{}}
    \toprule
    \textbf{Layer} & \textbf{Submodule} & \textbf{Total} & \textbf{Random} & \textbf{Subnet} & \textbf{Top-K (Ideal)} \\
    \midrule
    \multirow{7}{*}{5}  & q\_proj   & $16.38$ & $1.02$ & $3.82$ & $6.72$ \\
    & k\_proj   & $14.02$ & $0.88$ & $3.18$ & $5.70$ \\
    & v\_proj   & $83.97$ & $5.25$ & $18.69$ & $30.59$ \\
    & o\_proj   & $93.18$ & $5.82$ & $19.20$ & $29.70$ \\
    & up\_proj  & $133.12$ & $8.32$ & $20.48$ & $33.02$ \\
    & down\_proj& $144.38$ & $9.02$ & $21.89$ & $37.12$ \\
    & gate\_proj& $101.89$ & $6.37$ & $16.51$ & $27.26$ \\
    \midrule
    \multirow{7}{*}{15}  & q\_proj   & $15.49$ & $0.97$ & $4.83$ & $6.62$ \\
    & k\_proj   & $12.29$ & $0.77$ & $3.22$ & $5.25$ \\
    & v\_proj   & $48.64$ & $3.04$ & $12.93$ & $16.90$ \\
    & o\_proj   & $48.38$ & $3.02$ & $12.29$ & $14.98$ \\
    & up\_proj  & $65.02$ & $4.06$ & $15.30$ & $20.86$ \\
    & down\_proj& $70.14$ & $4.38$ & $15.42$ & $21.63$ \\
    & gate\_proj& $54.53$ & $3.41$ & $13.63$ & $19.07$ \\
    \midrule
    \multirow{7}{*}{25} & q\_proj   &  $6.94$ & $0.43$ & $2.82$ & $4.13$ \\
    & k\_proj   &  $6.02$ & $0.38$ & $1.90$ & $3.54$ \\
    & v\_proj   & $12.93$ & $0.81$ & $4.10$ & $5.98$ \\
    & o\_proj   & $13.38$ & $0.84$ & $4.61$ & $5.86$ \\
    & up\_proj  & $32.51$ & $2.03$ & $8.13$ & $10.82$ \\
    & down\_proj& $36.10$ & $2.26$ & $8.64$ & $11.20$ \\
    & gate\_proj& $25.09$ & $1.57$ & $6.50$ & $8.70$ \\
    \bottomrule
  \end{tabular}
  }
  \vspace{-10pt}
\end{table}

\subsubsection{Gradient- or Sensitivity-Based}
\label{sec:score}
\begin{figure}[H]
\centering
\includegraphics[width=0.9\columnwidth]{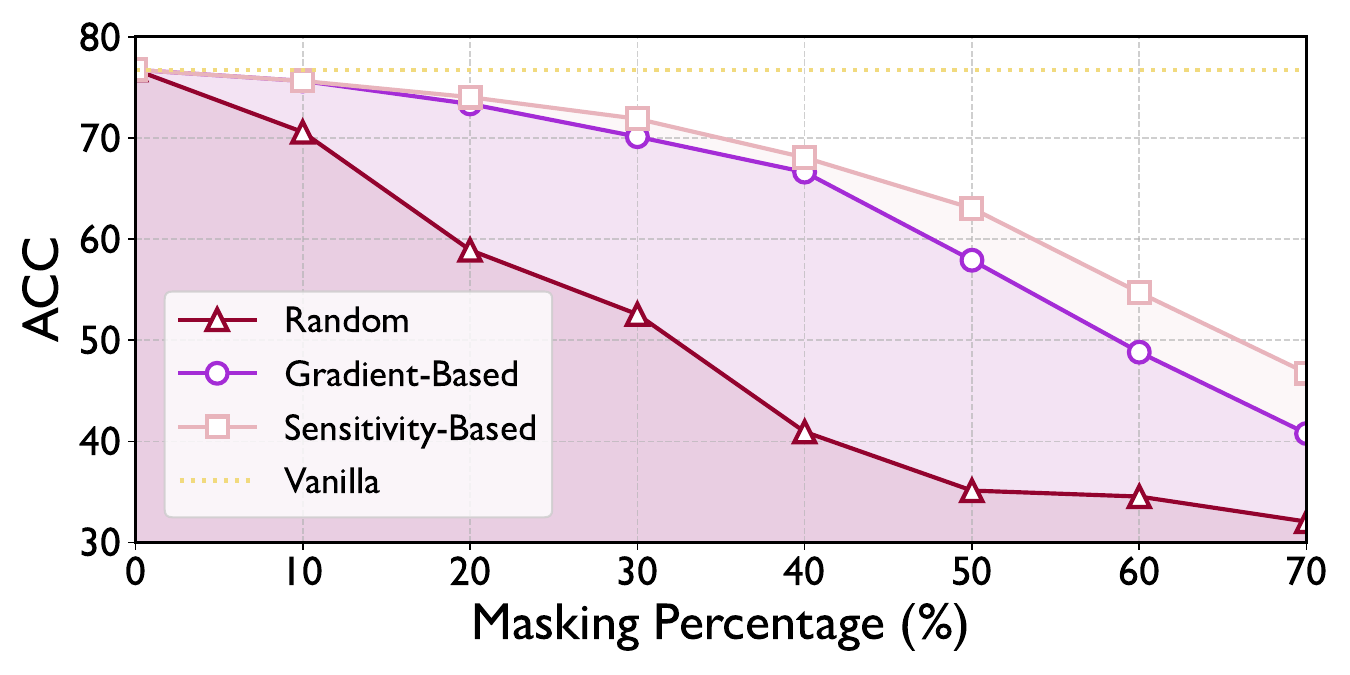}
\caption{ARC-E Accuracy under Different Masking Percentage. Linear layers in the 10-th to the 25-th decoder layer of LLaMA-2 7B are masked with gradient-based and sensitivity-based subnet selection strategies.}
\label{fig:vs}
\vspace{-10pt}
\end{figure}

Figure \ref{fig:vs} presents the performance on ARC-E across varying masking percentages. The gradient-based approach identifies the subnet based on magnitude of gradients while masking the remaining parameters. Among importance scoring strategies, the sensitivity-based approach, which is adopted by LoSiA, exhibits stronger robustness in higher masking ratios. However, tuning hyperparameters $\beta_1,\beta_2$ in the EMA of sensitivity-based importance calculation may result in marginal return for LoSiA, as evidenced by the minimal performance gap between the refined and unrefined selection methods.

\subsection{Experiments Details}
\label{sec:exp}
\subsubsection{Domain Specific Tasks}
 We randomly sample 50K data from open-source training datasets: MetaMathQA \cite{yu2023metamath}, Magicoder \cite{wei2023magicoder} and Alpaca-GPT4 \cite{peng2023instruction}, and evaluate fine-tuned models on GSM8K \cite{cobbe2021gsm8k}, MBPP \cite{austin2021program} and MMLU \cite{hendryckstest2021,hendrycks2021ethics}, respectively. Evaluations are conducted using lm-evaluation-harness \cite{eval-harness}, with baseline implementations from LLaMA-Factory \cite{zheng2024llamafactory}.

Table \ref{tab:hcdm} shows the hyperparameters for fine-tuning LLaMA-2 7B on MetaMathQA. We follow the commonly used configurations for baselines, while aligning GPU memory consumptions. For LoSiA, the hyperparameters for each task and model are listed in Table \ref{tab:hldm}. Rank factor $p$ is set to $\frac{1}{8}$, and the gradient dimension of lm\_head is compressed to a fraction by $p_o$. Time slot $T$ and learning rate may various across tasks. All experiments are conducted with single run on a NVIDIA A800-80GB GPU and CentOS 7 on x86-64 CPUs. Pytorch version is 2.4.1.
\begin{table*}[htb]
\centering
\caption{Hyperparameter Configurations of Fine-Tuning LLaMA-2 7B on MetaMathQA. Note that $\beta_1,\beta_2$ are EMA smoothing factors in sensitivity-based importance calculation, and are fixed across all experiments. $p$ and $p_o$ are dimension factors determining the shape of core subnets. $T$ of LoSiA refers to the time slot between re-selections.}
\label{tab:hcdm}
\renewcommand{\arraystretch}{1.4}
\resizebox{0.8\textwidth}{!}{%
\begin{tabular}{lcccc}
\specialrule{1pt}{0pt}{0pt}
& \textbf{LoRA/DoRA} & \textbf{PiSSA} & \textbf{GaLore} & \textbf{LoSiA}\\ \hline
Optimizer & \multicolumn{4}{c}{AdamW} \\
Epochs & \multicolumn{4}{c}{3} \\ 
Batch Size & \multicolumn{4}{c}{4} \\
LR & $2e-4$ & $1e-4$ & $1e-4$ & $6e-5$\\
Cutoff Length & \multicolumn{4}{c}{2048} \\
Warm-up Ratio & \multicolumn{4}{c}{0.1} \\ \hline
Rank Related & \multicolumn{2}{c}{$r=64$} & $r=512$ & $p=\frac{1}{8},\ p_o=\frac{1}{8}$\\
Scale Related & \multicolumn{2}{c}{$\alpha=128$} & $\alpha=2.0$ & - \\
Period Related & \multicolumn{2}{c}{-} & $T=200$ & $T=100$ \\ 
Others & \multicolumn{2}{c}{-} & Full Proj & $\beta_1=\beta_2=0.85$ \\
\hline
Implement Layer & \multicolumn{2}{c}{\makecell{proj\_q,proj\_k,proj\_v,proj\_o,\\up\_proj,down\_proj,gate\_proj}} & \multicolumn{2}{c}{\makecell{proj\_q,proj\_k,proj\_v,proj\_o,\\up\_proj,down\_proj,gate\_proj,\\lm\_head}} \\
\specialrule{1pt}{0pt}{0pt}
\end{tabular}
}
\end{table*}

\begin{table*}[htb]
\centering
\caption{Hyperparameter Configurations of LoSiA across different tasks and models.}
\label{tab:hldm}
\renewcommand{\arraystretch}{1.35}
\resizebox{0.8\textwidth}{!}{%
\begin{tabular}{lccc}
\specialrule{1pt}{0pt}{0pt}
\textbf{Datasets} & \textbf{MetaMathQA} & \textbf{Magicoder} & \textbf{Alpaca-GPT4} \\ \hline
LR & $6e-5$ & $5e-5$ & $5e-5$ \\
Time Slot $T$ & $100$ & $150$ & $150$ \\
Rank Factor $p$ & \multicolumn{3}{c}{$\frac{1}{8}$} \\ 
\addlinespace[0.3em]
\specialrule{0.5pt}{0pt}{0pt}
\textbf{Models} & \textbf{Gemma-2B} & \textbf{LLaMA-2 7B} & \textbf{LLaMA-2 13B} \\ \hline
Vocabulary Size & $256,000$ & $32,000$ & $32,000$ \\
Dimension Factor $p_o$ & $\frac{1}{64}$ & $\frac{1}{8}$ & 1 \\ 
\addlinespace[0.3em]
\specialrule{1pt}{0pt}{0pt}
\end{tabular}
}
\end{table*}

\subsubsection{Common-Sense Reasoning Tasks}
\begin{table}[H]
\centering
\caption{Datasets of Common-Sense Reasoning.}
\label{tab:dcr}
\renewcommand{\arraystretch}{1.4}
\resizebox{1.0\columnwidth}{!}{%
\begin{tabular}{lccc}
\specialrule{1pt}{0pt}{0pt}
\textbf{Datasets} & \textbf{\#Train} & \textbf{\#Test} & \textbf{Task Type} \\ \hline
\textbf{ARC-C} \cite{allenai:arc} & 1,120 & 1,170 & Q \& A\\
\textbf{ARC-E} \cite{allenai:arc} & 2,250 & 2,380 & Q \& A\\
\textbf{HellaSwag} \cite{zellers2019hellaswag} & 39,905 & 10,042 & Sentence Completion\\
\textbf{Winogrande} \cite{sakaguchi2019winograndeadversarialwinogradschema} & 9,248 & 1,267 & Fill the Blank \\
\textbf{PIQA} \cite{Bisk2020} & 16,100 & 1,840 & Q \& A \\
\textbf{OBQA} \cite{OpenBookQA2018} & 4,957 & 500 & Q \& A\\
\textbf{SIQA} \cite{sap2019socialiqacommonsensereasoningsocial} & 33,410 & 1,954 & Q \& A\\
\textbf{BoolQ} \cite{clark2019boolq} & 9,427 & 3,270 & Text Classification \\
\specialrule{1pt}{0pt}{0pt}
\end{tabular}
}
\end{table}

The datasets of common-sense reasoning tasks are presented in Table \ref{tab:dcr}, while corresponding hyperparameters detailed in Table \ref{tab:hcr}. The GPU memory usage remains aligned. For each PEFT baselines, searches in learning rate are performed. 

We report the accuracy metric evaluated by lm-evaluation-harness, which selects answers based on minimal perplexity. This approach mitigates the sensitivity of models to input phrasing variants, thereby enabling a more reliable measurement of the implicit knowledge encoded within the models.

\begin{table*}[htb]
\centering
\caption{Hyperparameter Configurations of Fine-Tuning LLaMA-2 7B on Common-Sense Reasoning Datasets.}
\label{tab:hcr}
\renewcommand{\arraystretch}{1.4}
\resizebox{0.8\textwidth}{!}{%
\begin{tabular}{lcccc}
\specialrule{1pt}{0pt}{0pt}
\hline
& \textbf{LoRA/DoRA} & \textbf{PiSSA} & \textbf{GaLore} & \textbf{LoSiA}\\ \hline
Optimizer & \multicolumn{4}{c}{AdamW} \\
Epochs & \multicolumn{4}{c}{3} \\ 
Batch Size & \multicolumn{4}{c}{16} \\
LR & $\{1e-4,2e-4\}$ & $\{5e-5,1e-4\}$ & $\{1e-4,2e-4\}$ & $\{5e-5,2e-4\}$\\
Cutoff Length & \multicolumn{4}{c}{256} \\
Warm-up Ratio & \multicolumn{4}{c}{0.1} \\ \hline
Rank Related & \multicolumn{2}{c}{$r=64$} & $r=512$ & $p=\frac{1}{8},\ p_o=1$\\
Scale Related & \multicolumn{2}{c}{$\alpha=128$} & $\alpha=2.0$ & - \\
Period Related & \multicolumn{2}{c}{-} & $T=200$ & $T=50$ \\ 
Others & \multicolumn{2}{c}{-} & Full Proj & $\beta_1=\beta_2=0.85$ \\
\hline
Implement Layer & \multicolumn{2}{c}{\makecell{proj\_q,proj\_k,proj\_v,proj\_o,\\up\_proj,down\_proj,gate\_proj}} & \multicolumn{2}{c}{\makecell{proj\_q,proj\_k,proj\_v,proj\_o,\\up\_proj,down\_proj,gate\_proj,\\lm\_head}} \\
\specialrule{1pt}{0pt}{0pt}
\end{tabular}
}
\end{table*}

\subsubsection{Rank Factor Robustness}
To evaluate the impact of the rank factor $p$, which determines the scale of selected core subnets, we conduct an ablation study on MetaMathQA as Table \ref{tab:lm} shows. Performance grows steadily with the number of training parameters on both Gemma 2B and LLaMA-2 7B. The results demonstrate LoSiA's robustness across various subnet scales. Note that $p=1/16$ may be relatively small for effective subnet fine-tuning, while increasing the computational budget boosts the performance. \looseness-1

\begin{table}[H]
\centering
\caption{Rank Factor Robustness of LoSiA on GSM8K}
\label{tab:lm}
\renewcommand{\arraystretch}{1.4}
\resizebox{0.8\columnwidth}{!}{%
\begin{tabular}{ccccc}
\specialrule{1pt}{0pt}{0pt}
\textbf{Model} & $\mathbf{1/16}$ & $\mathbf{1/8}$ & $\mathbf{1/4}$ & $\mathbf{1/2}$ \\ \hline
Gemma 2B & $37.53$ & $42.84$ & $45.03$ & $45.64$ \\
LLaMA-2 7B & $40.64$ & $44.66$ & $46.02$ & $48.45$ \\
\specialrule{1pt}{0pt}{0pt}
\end{tabular}
}
\end{table}

\begin{table}[htbp]
\centering
\renewcommand{\arraystretch}{1.05}
\caption{The Detail of Ablation Study on MMLU. Note that the variant \textit{GL} surpasses LoSiA on Humanities but shows performance drop on the rest of domains.}
\label{tab:glsl}
\resizebox{\columnwidth}{!}{%
\begin{tabular}{ccccccccccc}
\toprule
\multirow{2}{*}{\textbf{Model}} & \multicolumn{5}{c}{\textbf{MMLU}} \\
& \textbf{Humanities} & \textbf{Other} & \textbf{Social S} & \textbf{STEM} & \textbf{Avg.} \\ \midrule \addlinespace[0.4em] 
\makecell{Sensitivity-based \\ Localization (LoSiA)} &  $41.70$ & $\mathbf{52.23}$ & $\mathbf{50.89}$ & $\mathbf{36.82}$ & $\mathbf{44.95}$ \\ \addlinespace[0.3em] \hline \addlinespace[0.3em]
\makecell{Gradient-based \\ Localization (\textit{GL)}} & $\mathbf{42.64}$ & $51.41$ & $50.62$ & $36.22$ & $44.88$  \\ \addlinespace[0.3em] \bottomrule
\end{tabular}
}
\vspace{-5pt}
\end{table}

\subsubsection{Continue Learning}
\label{sec:cl}
To examine whether reduction of intruder dimensions in LoSiA mitigates forgetting in continue learning, we sequentially adapt LLaMA-2 7B through five common-sense reasoning tasks: HellaSwag, PIQA, BoolQ, SIQA and Winogrande. Learning rate for LoRA is $1e-4$ and for LoSiA is $5e-5$. The remaining hyperparameters are consistent with Table \ref{tab:hcr}. LoRA modules are merged into the backbone before subsequent task adaptation. 

Table \ref{tab:cldt} shows the detailed result during sequential adaptation. After continuing learning through all tasks, Seq-LoSiA outperforms Seq-LoRA across all benchmarks, highlighting its efficiency in forgetting mitigating.

\begin{table*}[htb]
\centering
\caption{Details of Performances on Continue Learning Five Common-Sense Reasoning Tasks. The column stands for training order, while the label "ST" indicates the result in single-tasking training.}
\label{tab:cldt}
\renewcommand{\arraystretch}{1.2}
\resizebox{0.9\textwidth}{!}{%
\begin{tabular}{lcccccc>{\columncolor{gray!30}}c}
\specialrule{1pt}{0pt}{0pt}
\textbf{Method} & \textbf{Task} & (\#1) HellaS & (\#2) PIQA & (\#3) BoolQ & (\#4) SIQA & (\#5) WinoG & ST \\ \hline
\multirow{5}{*}{\textit{Seq-LoRA}} & HellaSwag & $59.86$ & $55.64$ & $59.10$ & $57.86$ & $54.36$ & $59.86$ \\
 & PIQA & $76.01$ & $80.52$ & $77.86$ & $78.73$ & $77.64$ & $79.33$ \\
 & BoolQ & $77.80$ & $73.27$ & $86.30$ & $80.12$ & $75.93$ & $88.07$ \\
 & SIQA & $45.80$ & $47.80$ & $45.85$ & $59.52$ & $46.11$ & $56.86$ \\
 & Winogrande & $64.25$ & $68.35$ & $68.82$ & $69.93$ & $79.08$ & $73.88$ \\
\specialrule{0.5pt}{0pt}{0pt}
\multirow{5}{*}{\textit{Seq-LoSiA}} & HellaSwag & $63.72$ & $61.89$ & $61.11$ & $60.37$ & $\mathbf{56.43}$ & $63.72$ \\
 & PIQA & $78.29$ & $79.49$ & $79.82$ & $79.38$ & $\mathbf{77.75}$ & $81.50$ \\
 & BoolQ & $77.52$ & $70.76$ & $83.24$ & $82.54$ & $\mathbf{81.99}$ & $84.13$ \\
 & SIQA & $47.80$ & $48.26$ & $48.26$ & $59.93$ & $\mathbf{56.04}$ & $61.05$ \\
 & Winogrande & $68.51$ & $67.88$ & $68.51$ & $71.82$ & $\mathbf{80.19}$ & $77.19$ \\
\specialrule{1pt}{0pt}{0pt}
\end{tabular}
}
\end{table*}

\subsection{Resources Measurement}
\begin{figure}[H]
    \centering
    \vspace{10pt}
    \includegraphics[width=1.0\linewidth]{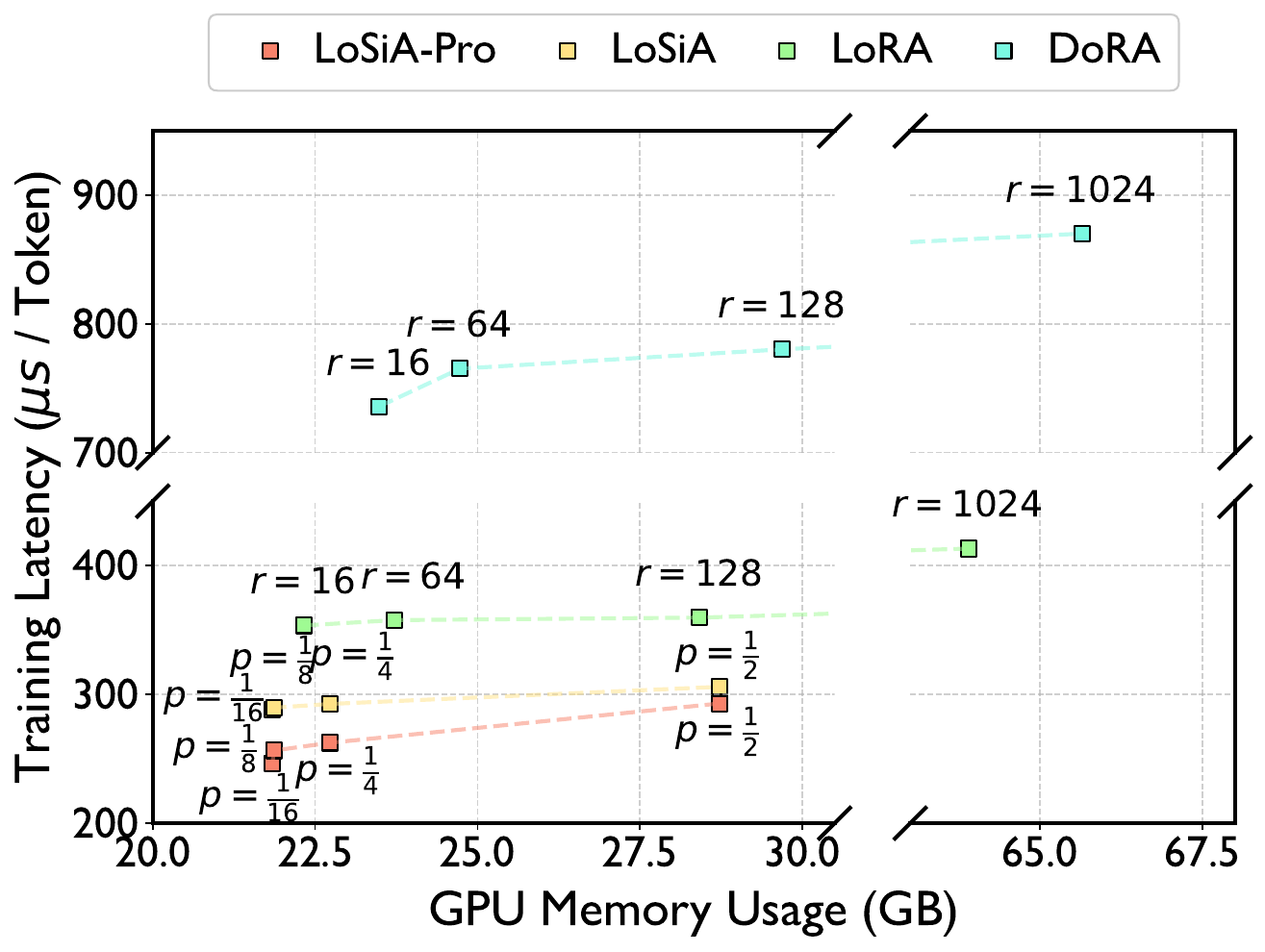}
    \caption{GPU Memory Usage and Training Latency Comparison \textsc{w Gradient Checkpointing}}
    \label{fig:oa}
\end{figure}

\begin{figure}[H]
    \vspace{10pt}
    \centering
    \includegraphics[width=1.0\linewidth]{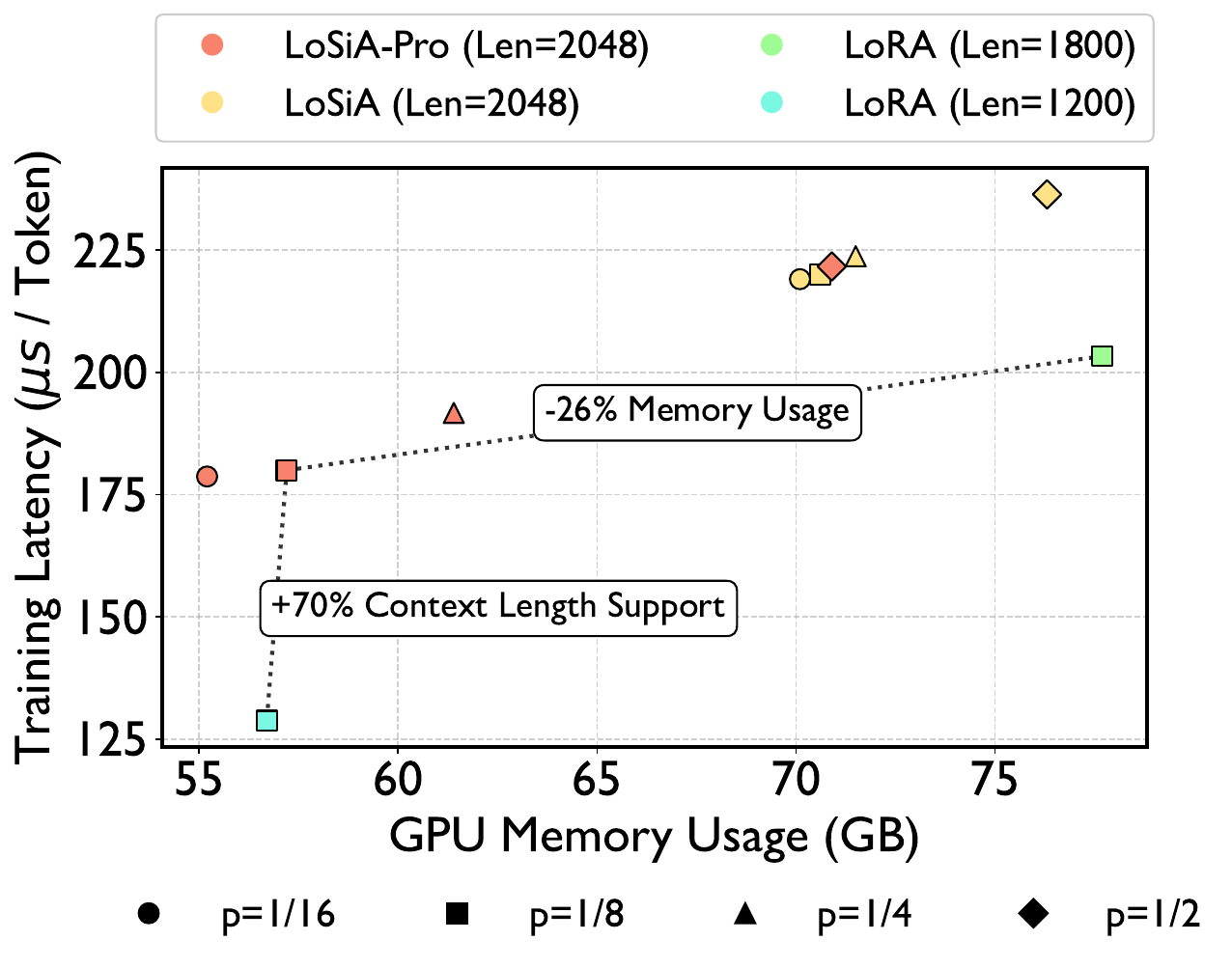}
    \caption{GPU Memory Usage and Training Latency Comparison \textsc{w/o Gradient Checkpointing}}
    \label{fig:oa2}
\end{figure}

Figure \ref{fig:oa} and \ref{fig:oa2} shows the memory and training time overheads for different PEFT methods on LLaMA-2 7B. With \textsc{Gradient checkpointing}, LoSiA and LoSiA-Pro display lower latency than low-rank methods across all ranks.

\label{sec:speed}

When disables \textsc{Gradient checkpointing}, LoSiA-Pro significantly reduces activation storage by at least $26\%$ and supports $70\%$ additonal training context length under consistent GPU memory constraints compared to LoRA.

\subsubsection{Memory Estimate}
Consider a model with $L$ decoder layers, each containing $K$ tunable matrices. The model use $b$-bit precision storage, with hidden dimension $d$ and vocabulary size $V$. Table \ref{tab:mm} shows GPU memory consumption details of LoRA, GaLore and LoSiA. For optimizers like AdamW, LoSiA reduces the gradient dimension of the output layer to a fraction $p_o$, while GaLore performs full fine-tuning on the output layer of shape $d\times V$. Both GaLore and LoSiA utilize per-layer weight update techniques for gradient computation. The update is therefore computed upon gradient acquisition and then promptly discarded.

In terms of auxiliary parameters, GaLore requires storing down- and up-projection matrices. Since $R$ is typically high-rank, GaLore's auxiliary parameters can be significantly larger than those of other methods, which may induce additional GPU memory consumptions.

For LoSiA, auxiliary parameters are used to compute the importance scores ($\overline{U}(\cdot)$ and $\overline{I}(\cdot)$). If gradient-based importance scoring is adopted, this component can be completely eliminated.

Regarding total memory consumption, increasing the rank in LoRA and GaLore incurs substantial overhead. However, in LoSiA, only the term $2(LKd^2p^2b + Vdp_ob)$ scales with rank factor $p$.
When using LoSiA-Pro without \textsc{Gradient Checkpointing}, only the activations corresponding to the input neurons need to be stored, making it the sole PEFT approach capable of mitigating this class of memory bottlenecks.

\begin{table}[H]
\centering
\caption{Comparison of Memory Consumptions. Cells in \textcolor{green!50!black}{green} highlight the components that may notably lower than other methods, while in \textcolor{red}{red} highlight the components that may cause relatively large memory consumption in high-rank.}
\label{tab:mm}
\renewcommand{\arraystretch}{1.4}
\resizebox{1.0\columnwidth}{!}{
\begin{tabular}{cccc}
\specialrule{1pt}{0pt}{0pt}
& LoRA & GaLore & \textbf{LoSiA} \\ \hline
\addlinespace[0.5em]
\textbf{Update Rank} & $r$ & $R$ & $pd$ \\

\addlinespace[0.5em]
\hline
\addlinespace[0.5em]
\textbf{\#Trainable} & $2LKrdb$ & $LKR^2b+Vdb$ & $LKd^2p^2b+$\textcolor{green!50!black}{$Vdp_ob$} \\
\textbf{\#Optimizer} & $4LKrdb$ & $2(LKR^2b+Vdb)$ & $2(LKd^2p^2b+$\textcolor{green!50!black}{$Vdp_ob$}) \\
\textbf{\#Gradient} & \textcolor{red}{$2LKrdb$} & \textcolor{green!50!black}{$\max\{d^2b,Vdb\}$} & \textcolor{green!50!black}{$\max\{d^2b,Vdb\}$} \\
\textbf{\#Auxiliary} & $2LKrdb$ & \textcolor{red}{$2LKRdb$} & $2Kd^2b$ \\ 
\addlinespace[0.5em]
\hline
\addlinespace[0.5em]
\textbf{\#Total} & $8LKrdb$ & \makecell{$2(LKR^2b+Vdb)$\\$+\max\{d^2b,Vdb\}$\\$+2LKRdb$} & \makecell{ $2(LKd^2p^2b+Vdp_ob)$\\$+\max\{d^2b,Vdb\}$\\$+2Kd^2b$} \\

\addlinespace[0.5em]
\specialrule{1pt}{0pt}{0pt}
\end{tabular}
}
\end{table}

\begin{table}[htb]
\centering
\caption{Details of Trainable Parameters for LoSiA under Different Hyperparameter Configurations on LLaMA-2 7B.}
\label{tab:tp}
\renewcommand{\arraystretch}{1.0}
\resizebox{1.0\columnwidth}{!}{%
\begin{tabular}{lcccc}
\specialrule{1.0pt}{0pt}{0pt}
\addlinespace[0.25em]
\rowcolor{gray!7}\multicolumn{5}{c}{\textit{\textbf{LoSiA}}} \\ 
\addlinespace[0.25em] \hline
$\mathbf{Factor\ p}$ & $1/16$ & $1/8$ & $1/4$ & $1/2$ \\ \hline
$\mathbf{Update\ Rank\ r}$ & $256$ & $512$ & $1024$ & $2048$ \\ 
\specialrule{0.8pt}{0pt}{0pt} \addlinespace[0.2em]
\rowcolor{gray!7}\multicolumn{5}{c}{$\mathbf{p_o=1/8}$} \\ \addlinespace[0.2em] \hline
$\mathbf{\#Trainable}$ & $42.8$M & $122.1$M & $439.3$M & $1700.8$M \\ \hline
$\mathbf{Mem(GB)}$ & $21.84$ & $21.87$ & $22.73$ & $28.73$ \\ \specialrule{0.8pt}{0pt}{0pt} \addlinespace[0.2em]
\rowcolor{gray!7}\multicolumn{5}{c}{$\mathbf{p_o=1}$} \\ \addlinespace[0.2em] \hline
$\mathbf{\#Trainable}$ & $158.0$M & $238.9$M & $562.2$M & $1855.7$M \\ \hline
$\mathbf{Mem(GB)}$ & $22.24$ & $22.84$ & $23.37$ & $28.98$ \\
\specialrule{1.0pt}{0pt}{0pt}
\end{tabular}
}
\end{table}

\subsubsection{Latency Measurement}

\begin{table}[htb]
\centering
\caption{Comparison of Training Latency on LLaMA-2 7B. Latencies are reported in measuerments of $\mu s$ per token, training with \textsc{Flash-Attention 2} \cite{dao2023flashattention2fasterattentionbetter}.}
\label{tab:training latency}
\renewcommand{\arraystretch}{1.4}
\resizebox{1.0\columnwidth}{!}{%
\begin{tabular}{ccccc}
\specialrule{1pt}{0pt}{0pt}
& \textbf{Forward} & \textbf{Backward} & \textbf{Other} & \textbf{Total} \\ \hline
\multicolumn{5}{c}{w \textsc{Gradient Checkpointing}}
\\ \hline

\addlinespace[0.5em]
LoRA$_{r=64}$ & 74.0 & 264.0 & 0 & 338.0 \\
DoRA$_{r=64}$ & 104.2 & 552.2 & 0 & 656.4 \\
GaLore$_{R=512}$ & \textbf{70.1} & 227.5 & \makecell{140.1 \\(574s / 500 step)} & 437.7 
\\ 
\addlinespace[0.5em]\hline

\addlinespace[0.5em]
\textbf{LoSiA}$_{p=\frac{1}{8}}$ & \textbf{70.0} (\textcolor{green!50!black}{-5.6\%}) & 220.4 (\textcolor{green!50!black}{-16.5\%}) & 0 & 290.4 (\textcolor{green!50!black}{-14.1\%})\\
\textbf{LoSiA-Pro}$_{p=\frac{1}{8}}$ & 71.4 (\textcolor{green!50!black}{-3.5\%}) & \textbf{173.4} (\textcolor{green!50!black}{-34.3\%}) & 0 & \textbf{244.8} (\textcolor{green!50!black}{-27.6\%})\\ 
\addlinespace[0.5em]\hline

\multicolumn{5}{c}{w/o \textsc{Gradient Checkpointing}} \\ \hline
LoRA$_{r=64}$ & \multicolumn{4}{c}{Out of Memory} \\
\textbf{LoSiA}$_{p=\frac{1}{8}}$ & \textbf{70.0} (\textcolor{green!50!black}{-5.6\%}) & 146.5 (\textcolor{green!50!black}{-44.5\%}) & 0 & 216.5 (\textcolor{green!50!black}{-35.1\%})\\
\textbf{LoSiA-Pro}$_{p=\frac{1}{8}}$ & 71.4 (\textcolor{green!50!black}{-3.5\%}) & \textbf{102.4} (\textcolor{green!50!black}{-61.3\%}) & 0 & \textbf{173.8} (\textcolor{green!50!black}{-49.6\%})
\\ 
\addlinespace[0.5em]
\specialrule{1pt}{0pt}{0pt}
\end{tabular}
}
\end{table}

We measure the training latency ($\mu s$ / token) fine-tuning with different PEFT methods on LLaMA-2 7B, and the results are shown in Table \ref{tab:training latency}. The experiments are conducted with $\text{cutoff\_len} =2048$ and $\text{batch
\_size}=4$. 

While demonstrating superior performance among existing baselines, LoSiA reduces training latency by 14.1\% compared to LoRA, 55.8\% compared to DoRA. The acceleration is mainly due to the elimination of low-rank matrix multiplication. Specifically, during backward propagation with \textsc{Gradient Checkpointing}, the production of low-rank matrices introduces significant overhead for activation recomputation and gradient calculation. Note that LoRA can avoid gradient calculations on backbone weights, but this requires specialized implementations and still introduces a large coefficient of computational complexity.

For LoSiA-Pro, the computational complexity remains the same as LoSiA during the forward pass, but it only requires storing a proportion $p$ of the input activations of the linear layers. During the backward pass, LoSiA-Pro reduces the computational cost to $p^2$  relative to full gradient computation, which significantly lowers the latency of backward propagation. This results in highly efficient training and lower GPU memory consumption.

\subsubsection{Algorithm}
The core of LoSiA is summarized in Algorithm \ref{alg:losia}. The model is partitioned into weight groups (in this paper, simply the decoder layers); each group is assigned its own LoSiA optimizer that receives the total number of groups ($L$) and related meta-information, and LoSiA automatically performs the weight updates during the backward pass.

\algnewcommand\UPDATE{\item[\textbf{UPDATE}]}
\algblockdefx[UPDATE]{UPDATE}{ENDUPDATE}%
  [1]{\textbf{WEIGHTS OPTIMIZATION BY \textsc{Adam}}}%
  {}
\begin{algorithm*}[ht]
\caption{Pseudo Code of LoSiA}
\label{alg:losia}
\begin{algorithmic}[1]
\Require
  Weight matrix $W\in \mathbb{R}^{n\times m}$ lying in decoder layer (weight groups) $l$; Total decoder layer (weight groups) $L$; 
  EMA ratio $\beta_1,\beta_2$; Adam decay rates $\beta'_1,\beta'_2$;
  Rank factor $p$; Time slot $T$;

\Statex
\State
  Initialize scales of the core subnet $n_p\gets \lfloor np\rfloor, m_p\gets \lfloor mp\rfloor$
\State
  Initialize first- and second-order momentum $M_{0}\gets 0_{n_p\times m_p}$,\quad
  $V_{0}\gets 0_{n_p\times m_p}$
\State
  Initialize selected neurons randomly $\rho\gets \text{random}([1\ldots n],n_p)$,\quad
  $\gamma\gets \text{random}([1\ldots m],m_p)$
\State
  Initialize training step $t\gets 1$
\Statex
\Repeat
  \State $t'\gets (t-1)\mod (TL)$
  \If{$\lfloor\frac{t'}{T}\rfloor=l-1$}  \Comment{\{sensitivity-based parameter importance estimation\}}
    \State $ I\gets W\cdot \nabla W$
    \State $ I\gets |I-\frac{1}{2}I^2|$  \Comment{\{calculation of importance score by Eq.\ref{eq:imp}\}}
    \Statex 
    \If{$t$ is the first step of time slot $T$}
      \State
        $\overline{I}_{t-1}\gets 0_{n\times m}$,\quad
        $\overline{U}_{t-1}\gets 0_{n\times m}$
    \EndIf
    \Statex \Comment{\{exponential moving average for importance and uncertainty\}}
    \State $ \overline{I}_t \gets \beta_1\overline{I}_{t-1}+(1-\beta_1)\overline{I}_t$ 
    \State $ \overline{U}_t \gets \beta_2\overline{U}_{t-1}+(1-\beta_2)\overline{U}_t$  
  \EndIf
  
\hrulefill
  \UPDATE {}
      \State ($\nabla W_{\rho,\gamma}\gets \tilde{L}_S\tilde{R}_S$) \Comment{\{partial activation $\tilde{L}_S$ manually saved during forward in LoSiA-Pro\}}
      \State $G\gets \nabla W_{\rho,\gamma}$ \Comment{\{obtain subnet gradient by indices selection\}}
      \State $M_t\gets \beta'_1 M_{t-1}+(1-\beta'_1)\cdot G$
      \State $V_t\gets \beta'_2 V_{t-1}+(1-\beta'_2)\cdot G^2$
      \State $M_t\gets M_t/(1-\beta'_1)$,\quad
             $V_t\gets V_t/(1-\beta'_2)$
      \State $N_t\gets M_t/(\sqrt{V_t}+\epsilon)$
      \Statex 
      \State $\eta \gets \overline{lr}(t)$ \Comment{\{calculate learning rate using rewarming scheduler in Section \ref{sec:rem}\}}
      \State $W_{t}\gets W_{t-1}-\eta N_t$
    \ENDUPDATE \hrulefill
    \Statex
  \If{$t'\mod T=0$ and $\frac{t'}{T}=l$} 
    \State $ s_W\gets \overline{I}_{t-1}\cdot \overline{U}_{t-1}$
    \For {localization algorithm $\mathcal{A}_i$}
        \State $\rho_i,\gamma_i \gets \mathcal{A}_i(s_W,p)$
    \EndFor
    \State $k=\arg\max_{i} S(s_{W_{\rho_i,\gamma_i}})$ \Comment{\{choose the core subnet with the highest importance\}}
    \State $\rho,\gamma \gets \rho_k,\gamma_k$
    \State $\text{delete}\ \overline{I},\overline{U}$ \Comment{\{delete useless tensors after importance calculation terminate\}}
    \State $M_{t}\gets 0_{n_p\times m_p},V_{t}\gets 0_{n_p\times m_p}$
  \EndIf
  \Statex 
  \State $t\gets t+1$
\Until{training finishes}
\State \Return $W$
\end{algorithmic}
\end{algorithm*}

\end{document}